%% file: camera_ready.tex
\documentclass[letterpaper]{article} 
\usepackage{aaai25}  
\usepackage{times}  
\usepackage{helvet}  
\usepackage{courier}  
\usepackage[hyphens]{url}  
\usepackage{graphicx} 
\usepackage{natbib}  
\usepackage{caption} 
\usepackage{algorithm}
\usepackage{algorithmic}

\usepackage{newfloat}
\usepackage{listings}
\DeclareCaptionStyle{ruled}{labelfont=normalfont,labelsep=colon,strut=off} 
\floatstyle{ruled}
\newfloat{listing}{tb}{lst}{}
\floatname{listing}{Listing}
\usepackage{xspace}
\makeatletter
\DeclareRobustCommand\onedot{\futurelet\@let@token\@onedot}
\def\@onedot{\ifx\@let@token.\else.\null\fi\xspace}
\def\eg{\emph{e.g}\onedot} 
\def\ie{\emph{i.e}\onedot}

\makeatother
\usepackage{color}
\usepackage{leftidx}
\usepackage{amsmath}
\usepackage{amssymb}
\usepackage{mathtools}
\usepackage{bbding}
\usepackage{booktabs}
\usepackage{multirow}
\usepackage{makecell}
\usepackage{enumitem}
\usepackage{dsfont}
\usepackage{array}

\newcommand{\ourdataset}[1]{HybridMatch}
\newcommand{\ourlodataset}[1]{HybridLoMatch}
\newcommand{\ourmethod}[1]{HybridReg}

\title{\ourmethod{}: Robust 3D Point Cloud Registration with Hybrid Motions}
\author{
    Keyu Du\textsuperscript{\rm 1}\equalcontrib, 
    Hao Xu\textsuperscript{\rm 2,3}\equalcontrib,
    Haipeng Li\textsuperscript{\rm 1,4},
    Hong Qu\textsuperscript{\rm 1},
    Chi-Wing Fu\textsuperscript{\rm 2,3},
    Shuaicheng Liu\textsuperscript{\rm 1,4}\thanks{Corresponding author.}
}
\affiliations{
    \textsuperscript{\rm 1}University of Electronic Science and Technology of China\\
    \textsuperscript{\rm 2}Department of Computer Science and Engineering, CUHK\\
    \textsuperscript{\rm 3}Institute of Medical Intelligence and XR, CUHK\\
    \textsuperscript{\rm 4}Megvii Technology \\
    \{dukeyu@std., lihaipeng@std., hongqu@, liushuaicheng@\}uestc.edu.cn,
    \{xuhao, cwfu\}@cse.cuhk.edu.hk
}

\usepackage{bibentry}

\begin{document}

\maketitle

\begin{abstract}
Scene-level point cloud registration is very challenging when considering dynamic foregrounds.
Existing indoor datasets mostly assume rigid motions, so the trained models cannot robustly handle scenes with non-rigid motions. 
On the other hand, non-rigid datasets are mainly object-level, so the trained models cannot generalize well to complex scenes. 
This paper presents \ourmethod{}, a new approach to 3D point cloud registration, learning uncertainty mask to account for hybrid motions: rigid for backgrounds and non-rigid/rigid for instance-level foregrounds. 
First, we build a scene-level 3D registration dataset, namely \ourdataset{}, designed specifically with strategies to arrange diverse deforming foregrounds in a controllable manner. 
Second, we account for different motion types and formulate a mask-learning module to alleviate the interference of deforming outliers. 
Third, we exploit a simple yet effective negative log-likelihood loss to adopt uncertainty to guide the feature extraction and correlation computation. 
To our best knowledge, \ourmethod{} is the first work that exploits hybrid motions for robust point cloud registration. 
Extensive experiments show \ourmethod{}'s strengths, leading it to achieve state-of-the-art performance on both widely-used indoor and outdoor datasets. 
\end{abstract}

%
\begin{links}
    \link{Code}{https://github.com/hxwork/HybridReg_PyTorch}
\end{links}

\begin{figure}[t]
    \centering
    \includegraphics[width=0.98\linewidth]{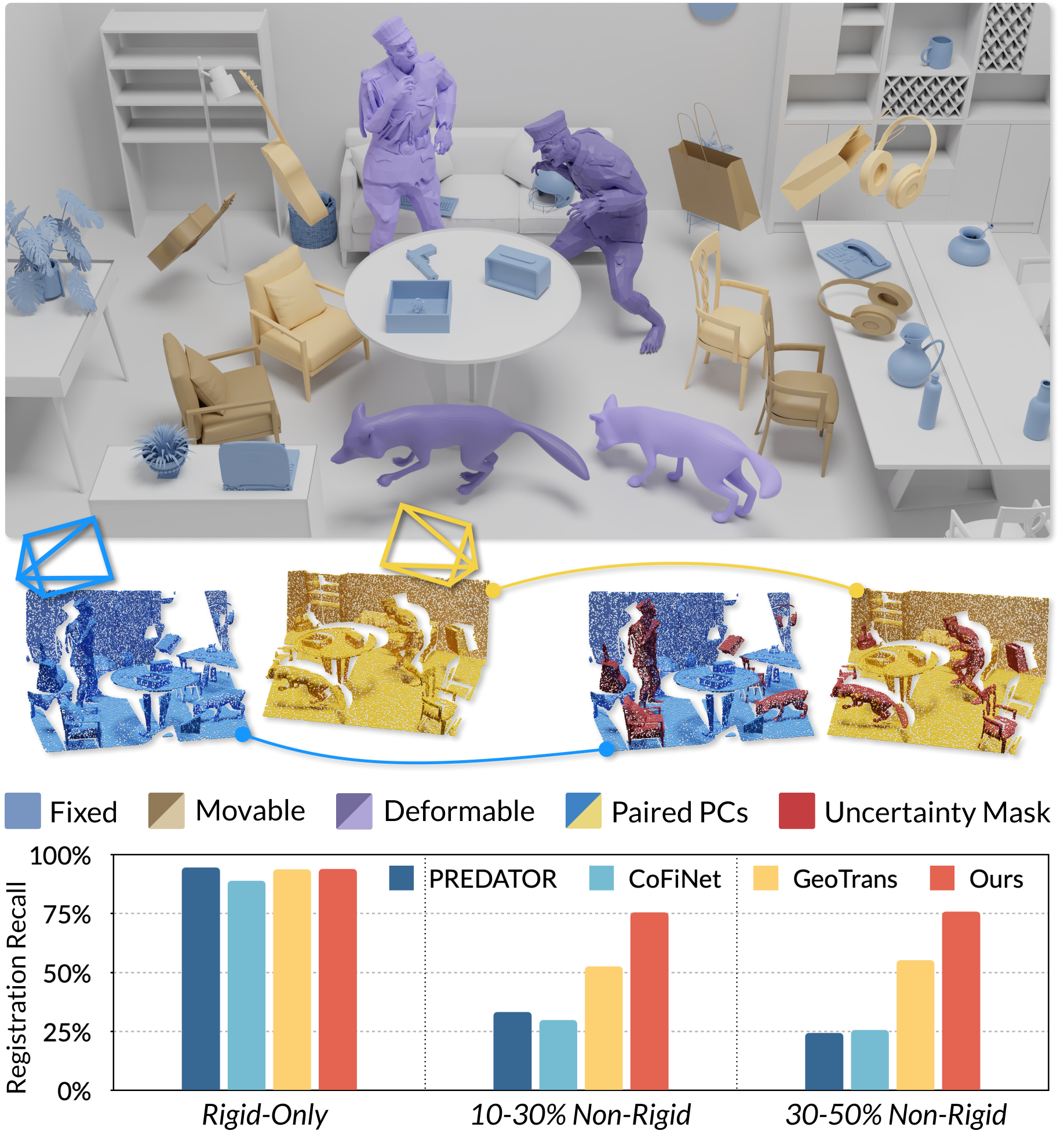}
    \caption{
    We build \ourdataset{}, the first large-scale indoor synthetic dataset with hybrid motions for 3D point cloud registration (top).
    To improve the robustness, we further design \ourmethod{}, integrating probabilistic uncertainty masks into correspondence estimation (middle), achieving significant boosts (below) in challenging hybrid-motion scenes.
    }
    \label{fig: teaser}
\end{figure}

\section{Introduction}

3D point cloud registration is vital for diverse applications, attracting incredible attention from industry and academia. 
Recent data-driven methods have shown remarkable success, particularly in indoor scenes~\cite{bai2020d3feat, huang2020predator, yu2021cofinet}, thanks to large-scale datasets like 3DMatch~\cite{zeng20173dmatch}, SIRA-PCR~\cite{Chen_2023_ICCV}, and PointRegGPT~\cite{chen2024pointreggpt}.

Typically, the input pair of point clouds are partially overlapping, with 
(i) rigid background motions due to the camera motion, and 
(ii) non-rigid foreground motion due to moving or deforming objects.
To align the background, algorithms should eliminate the foreground interference and determine a 3D rigid transformation.
However, most existing indoor datasets assume rigid motions in the whole scene, {\em neglecting\/} common non-rigid foreground motions.
So, the first challenge is the lack of a comprehensive benchmark for model training and evaluation on scenes with hybrid motions.

Though directly training on real-captured datasets can alleviate this issue, data collection is highly time-consuming, and data annotation is expensive and error-prone.
An appealing alternative to overcome the data scarcity is to leverage simulated data,~\eg, Lepard~\cite{lepard2021} introduces 4DMatch, an object-level synthetic dataset for non-rigid point cloud registration. 
Yet, models trained on object-level data have proven to struggle to generalize to complex indoor scenes~\cite{qin2022geometric, huang2020predator}, as objects are simpler and have strong shape priors.
To sum up, the second challenge is how to design a scene-level benchmark for 3D point cloud registration with hybrid motions, to efficiently account for both data collection and annotations.

Further, even with such appropriate data, algorithms still need to handle outliers in non-overlapping regions, caused not only by different captured views but also by the dynamic foregrounds.
While powerful feature descriptors help address the former, they lack specific designs to handle the latter, as foreground points with similar local geometric structures may be outliers due to inconsistent motions with the background. 
Directly applying prior methods to hybrid-motion scenes leads to heavy performance degradation.
Hence, the third challenge is how to integrate feature extraction and correspondence matching with precise guidance to account for hybrid motions.

In this work, to address the first two challenges, we draw inspiration from FlowNet~\cite{dosovitskiy2015flownet} and SIRA-PCR~\cite{Chen_2023_ICCV} to create the first large-scale indoor synthetic dataset with hybrid motions, named \ourdataset{}. Specifically, beyond the rigid backgrounds provided by 3D-FRONT~\cite{fu20213d}, we simulate non-rigid motions by applying instance-level rigid motions to objects from ShapeNet~\cite{chang2015shapenet} on the surfaces of some furniture and in mid-air, combined with deforming objects from DeformingThings4D~\cite{li20214dcomplete}.

For the third challenge, training existing models on simulated point cloud pairs proves insufficient due to the increased difficulty in correspondence matching caused by hybrid motions. As shown in Fig.~\ref{fig: teaser}, their performances degrade as the ratio of non-rigid motion increases. To address it, inspired by the 2D motion estimation method~\cite{truong2023pdc}, we propose a probabilistic model named \textbf{\ourmethod{}} to learn uncertainty masks that address \textbf{Hybrid} motions in robust 3D point cloud \textbf{Reg}istration.
Rather than deterministically predicting correspondences, our reformulated probabilistic model computes the correspondence relationships while simultaneously learning a robust confidence map to model correspondence-wise uncertainty. This map indicates the reliability of the correspondence predictions, which is crucial for accurately solving the 3D transformation. To facilitate training, we assume the output follows a Laplace distribution and employ a simple yet effective negative log-likelihood (NLL) loss to incorporate probabilistic uncertainty into the loss function, enhancing robust correspondence matching. Our main contributions are three-fold:
\begin{itemize}
\item
We construct \ourdataset{}, the first large-scale indoor synthetic dataset with hybrid and diverse motions for 3D point cloud registration. 
\item
We design \ourmethod{}, a probabilistic model to learn uncertainty masks, guided by an effective negative log-likelihood loss, to account for hybrid motions and enhance feature extraction and correspondence matching.
\item
Extensive qualitative and quantitative comparisons on both widely-used indoor and outdoor datasets demonstrate the state-of-the-art performance of our approach.
\end{itemize}

\section{Related Work}
\paragraph{Rigid 3D point cloud registration.}
%
Most approaches are correspondence-based, evolved from traditional handcrafted features~\cite{besl1992method, rusinkiewicz-normal-sampling, yang2013go, rusinkiewicz2019symmetric, FPFH} and RANSAC~\cite{fischler1981random} to deep-learning-based feature descriptors~\cite{choy2019fully, gojcic2019perfect, bai2020d3feat, ao2021spinnet, poiesi2022learning, wang2022you, yu2021cofinet, yew2022regtr, wang2023roreg, yangone, lepard2021, Ao_2023_CVPR, yu2023peal,  Mei_2023_CVPR, Liu_2023_ICCV, Chen_2023_ICCV, liu2023regformer}, including weighted SVD for efficient transformation derivation~\cite{wang2019deep, wang2019prnet, idam, yew2020-RPMNet, fu2021robust} and deep robust estimators for transformation accuracy~\cite{bai2021pointdsc, choy2020deep, pais20203dregnet, Jiang_2023_CVPR, Zhang_2023_CVPR, Hatem_2023_ICCV}. 
Others are correspondence-free, directly regressing transformations,~\eg,~\cite{aoki2019pointnetlk, xu2021omnet, xu2022finet}, yet they struggle with scalability and generalization when handling complex scenes~\cite{qin2022geometric}.
Existing methods are mainly designed for (i) indoor scenarios, such as 3DMatch~\cite{zeng20173dmatch}, 7Scenes~\cite{shotton2013scene}, SIRA-PCR~\cite{chen2023sira}, and PointRegGPR~\cite{chen2024pointreggpt}, or (ii) object-level scenarios,~\eg, ModelNet40~\cite{chang2015shapenet}, Stanford 3D Scan~\cite{curless1996volumetric}, and AutoSynth~\cite{Dang_2023_ICCV}. 
All of them, however, consider only rigid motions, ignoring the commonly existing non-rigid motions in practice, \eg, moving humans/objects.
In this work, we create the first large-scale indoor synthetic dataset with hybrid motions, enabling training and testing to effectively account for deformable foregrounds.

\begin{figure*}[t]
    \centering
    \includegraphics[width=0.995\linewidth]{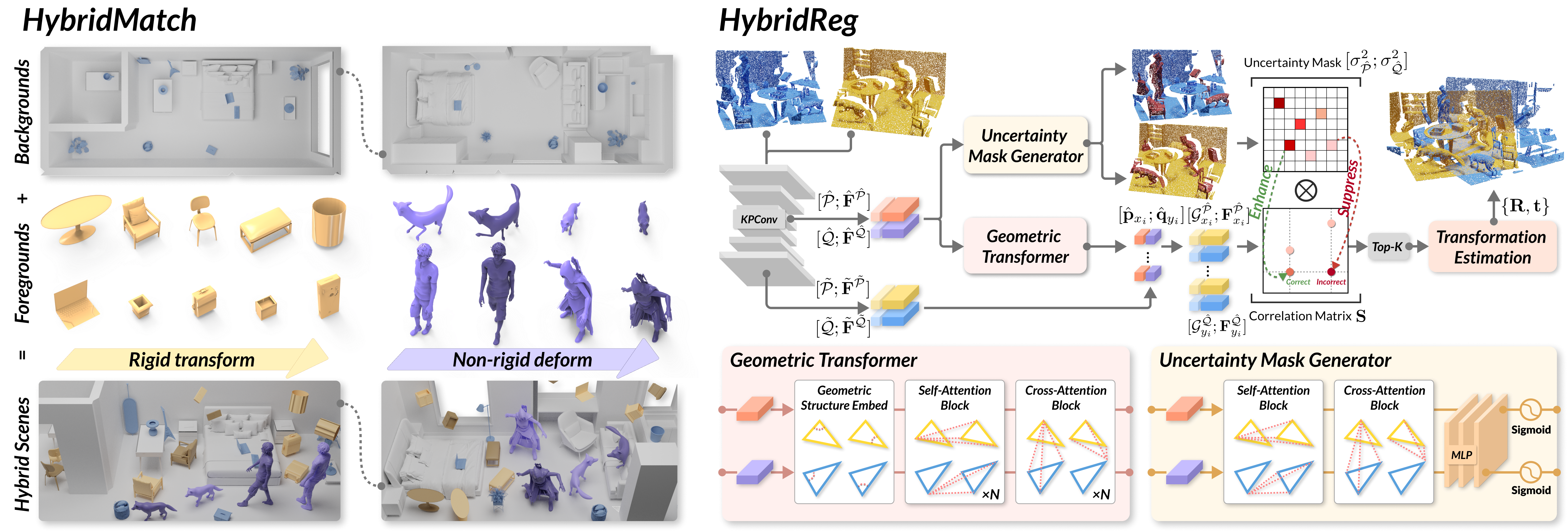}
    \caption{Our framework: (i) \ourdataset{} is a large-scale indoor synthetic dataset with hybrid motions for point cloud registration, containing diverse movable and deformable foreground objects; and (ii) \ourmethod{} is designed to estimate uncertainty mask for enhancing discriminative feature extraction and accurate correspondence matching against hybrid motions.
    }
    \label{fig: pipeline}
\end{figure*}

\paragraph{Non-rigid correspondence estimation} is a major topic in dynamic reconstruction~\cite{bozic2020deepdeform, gao18surfelwarp, newcombe2015dynamicfusion} and geometry processing~\cite{huang2008non, groueix20183d, ovsjanikov2012functional}, focusing on estimating non-rigid correspondences from deformable point clouds or manifold surfaces.
DynamicFusion~\cite{newcombe2015dynamicfusion} employs projective correspondence for real-time efficiency and is further utilized for dense correspondence learning in~\cite{schmidt2016self}.
VolumeDeform~\cite{innmann2016volumedeform} adopts SIFT~\cite{lowe2004distinctive} for robust non-rigid tracking. 
DeepDeform~\cite{bozic2020deepdeform} learns sparse global correspondence for patches in deforming RGB-D sequences. 
\cite{li2020learning} learn non-rigid features via a differentiable non-rigid alignment optimization. 
NNRT~\cite{bozic2020neural} estimates correspondence end-to-end with an outlier rejection network. 
Besides, scene flow estimation,~\eg,~\cite{li2021neural, liu2019flownet3d, wu2020pointpwc}, is a closely-related technique that delivers point-wise correspondence.
The key difference between our task and these works is that non-rigid correspondences on moving or deforming foreground objects are outliers and should be filtered out for accurate background alignment in our scene-level task.

\section{Method}

\subsection{Overview}
Fig.~\ref{fig: pipeline} gives an overview of our approach, which includes
(i) \ourdataset{}, a large-scale indoor synthetic dataset with hybrid motions for 3D point cloud registration (Sec.~\ref{sec: 3.2});
(ii) \ourmethod{}, the method we designed to estimate probabilistic uncertainty masks and evaluate reliability and accuracy of the correspondence prediction against hybrid motions (Sec.~\ref{sec: 3.3}); and
(iii) our loss function design (Sec.~\ref{sec: 3.4}).

\subsection{\ourdataset{} Dataset} 
\label{sec: 3.2}
Existing indoor datasets typically contain only static scenes, yet dynamic scenes are common in the real world. To better simulate real-world environments, our \ourdataset{} includes not only rigid background motions but also a diverse range of rigid and non-rigid foreground motions for objects within the background scene, such as humans moving around, animals wandering, and objects or furniture shifting. This makes \ourdataset{} more representative of the complexities found in everyday indoor environments.
To achieve this level of realism, we carefully design different strategies for various types of motion.

\paragraph{Rigid scene-level background motions.}
We construct the indoor scene background using 3D-FRONT, as it offers professionally-designed room layouts and a rich variety of furniture.
To enhance the scene complexity, following~\cite{chen2023sira}, models from the ShapeNet dataset are randomly scaled, rotated, and placed either mid-air in the rooms or on existing furniture items such as tables and beds.
The rigid background motion is naturally introduced when adjusting the camera view in the point cloud rendering process.

\paragraph{Non-rigid/rigid object-level foreground motions.}
Given a room layout, we consider two distinct cases of injecting foreground object motions based on the rigidity of the object:
(i) Deformable non-rigid objects, such as humans and animals. 
Inspired by~\cite{lepard2021}, we employ models and offsets from the DeformingThings4D dataset, a synthetic collection, comprising thousands of animation sequences across various categories of humanoids and animals. 
For each scene, animation sequences are randomly selected; half feature a single human and the other half feature multiple humans and animals.
To enhance the diversity in movement magnitude, we employ random offsets to the human and animal models in every frame.
(ii) Movable rigid objects, such as chairs and cups. 
First, additional rigid models from ShapeNet are positioned around each deformable non-rigid object in (i), with random XY-plane offsets applied to each frame to simulate movement and introduce noise.
To augment realism, we introduce subtle movements to movable furniture items, thereby enriching the scene's plausibility with hybrid motions.
Care is taken to ensure that each foreground object insertion is collision-free, maintaining the physical plausibility of the scene.

\paragraph{Registration data preparation.}
To produce paired data for point cloud registration, we follow~\cite{chen2023sira} to use a virtual perspective camera to render a depth map for each viewpoint, subsequently converting these maps into point clouds. 
We limit the depth range and randomly remove half of the large flat surfaces, such as walls and floors, to emphasize the foreground. 
To ensure scene complexity, high-quality camera views are selected. 
The viewing range is set to [0\textdegree,360\textdegree] horizontally and [0\textdegree,45\textdegree] vertically, with views uniformly sampled at intervals of 30\textdegree and 15\textdegree, respectively.
Ground-truth correspondences of the overlapping point cloud pairs are easily established, given the accurate camera poses. 
Following 3DMatch, we categorize our data into \ourdataset{} ($>$30\% overlap ratio) and \ourlodataset{} (10-30\% overlap ratio).
Based on the proportion of non-rigid motions, each set is further divided into \emph{10-30\% non-rigid} and \emph{30-50\% non-rigid} splits.
To validate our data, each pair is accompanied by a corresponding pair without any animations, forming a \emph{rigid-only} split.
For each split, we create a validation/test set, consisting of 100/1,000 pairs.

\subsection{\ourmethod{}} 
\label{sec: 3.3}
\paragraph{Problem formulation.}
Given the input point cloud pair $\mathcal{I} = (\mathcal{P}, \mathcal{Q})$, where $\mathcal{P} \!\in\! \mathbb{R}^{N \times 3}$ is the source point point with $N$ points and $\mathcal{Q} \!\in\! \mathbb{R}^{M \times 3}$ is the target with $M$ points.
A network $F$ with parameters $\theta$ is employed to estimate the correspondences $\mathcal{C}$, \ie, $\mathcal{C} \!=\! F(\mathcal{I};\theta)$. 
Then, the 3D transformation $\{\mathbf{R}\!\in\! S O(3),\mathbf{t} \!\in\! \mathbb{R}^3\}$ is derived from $\mathcal{C}$.

\paragraph{Feature extraction.}
We first downsample the source and target points to obtain superpoints $\hat{\mathcal{P}}$ and $\hat{\mathrm{Q}}$ as in~\cite{qin2022geometric}. 
Utilizing KPConv-FPN~\cite{thomas2019kpconv}, we extract features of the point clouds, with the associated learned features denoted as $\mathbf{F}^{\hat{\mathcal{P}}} \!\in\! \mathbb{R}^{\hat{N} \times d}$ and $\mathbf{F}^{\hat{\mathcal{Q}}} \!\in\! \mathbb{R}^{\hat{M} \times d}$. 

\paragraph{Geometric transformer.}
Following~\cite{qin2022geometric}, we employ geometric self-attention mechanisms to learn global correlations among superpoints within each point cloud. 
Here, we describe the computation for $\hat{\mathcal{P}}$, and the same process applies to $\hat{\mathcal{Q}}$. 
Given the input feature matrix $\mathbf{X} \!\in\! \mathbb{R}^{\hat{N}\times{d_t}}$, the self-attention feature $\mathbf{Z} \!\in\! \mathbb{R}^{\hat{N}\times{d_t}}$ is computed as the cumulative weighted sum of all projected input features: 
$\mathbf{Z}_i\!=\!\sum\nolimits_{j=1}^{\hat{N}}a_{i,j}({\mathbf{X}_j}\mathbf{W}^V)$,
where the weight coefficient $a_{i,j}\!=\!\operatorname{Softmax}(\frac{(\mathbf{X}_i\mathbf{W}^Q)(\mathbf{X}_j\mathbf{W}^K +\mathbf{R}_{i,j}\mathbf{W}^R)^T}{\sqrt{d_t}})$. 
$\mathbf{R}_{i,j} \!\in\! \mathbb{R}^{d_t}$ represents the geometric structure embedding, which includes pairwise distances and angular information among points.
$\mathbf{W}$ denotes the projection matrix, whose superscripts $Q$, $K$, $V$, and $R$ correspond to queries, keys, values, and geometric structure embeddings, respectively.
Then, to encode inter-frame geometric consistency, cross-attention mechanisms are adopted.
The cross-attention feature $\mathbf{Z}^{\hat{\mathcal{P}}} \!\in\! \mathbb{R}^{\hat{N}\times{d_t}}$ for $\hat{\mathcal{P}}$ is computed as 
$\mathbf{Z}_i^{\hat{\mathcal{P}}}\!=\!\sum\nolimits_{j=1}^{\hat{M}}a_{i,j}({\mathbf{X}_j^{\hat{\mathcal{Q}}}}\mathbf{W}^V)$,
where the weight coefficient $a_{i,j}\!=\!\operatorname{Softmax}(\frac{(\mathbf{X}_i^{\hat{\mathcal{P}}}\mathbf{W}^Q)(\mathbf{X}_j^{\hat{\mathcal{Q}}}\mathbf{W}^K)^T}{\sqrt{d_t}})$.

\paragraph{Uncertainty mask generation.}
Previous methods typically predict correspondences $\mathcal{C}$ directly by $\mathcal{C} \!=\!\! F(\mathcal{I}; \theta)$, however, it does not provide confidence measures.
Instead, our objective is to estimate the conditional probability density function $p(\mathcal{C} | \mathcal{I};\theta)$, which allows us to capture the uncertainty associated with the predictions. 
To achieve this, we design a network to predict the parameters $\Phi(\mathcal{I};\theta)$ of a family of distributions, \ie,
\begin{equation}
    p(\mathcal{C} | \mathcal{I};\theta)=p(\mathcal{C} | \Phi(\mathcal{I};\theta))=\Pi_{i,j}p(c_{i,j} | \varphi_{i,j}(\mathcal{C;\theta})).
\end{equation}
For tractable estimation, we assume conditional independence between the predictions for the $i$-th source point and the $j$-th target point.
Here, $c_{i,j} \!\in\! \{0,1\}$ is the ground-truth correspondence relationship, and $\varphi_{i,j} \!\in\! \mathbb{R}^n$ denotes the predicted parameters, including both the predicted correspondence relationship and the confidence. 
For simplicity, we omit the subscript $i,j$ in the subsequent discussion.

Compared to the direct prediction approach $\mathcal{C} \!=\! F(\mathcal{I}; \theta)$, the parameters $\Phi(\mathcal{I}; \theta)$ of the predictive distribution encode additional information about the correspondence, including its uncertainty. 
Following previous works~\cite{ilg2018uncertainty, kendall2017uncertainties, truong2023pdc}, this is done by predicting the variance of the correspondence. 
In these cases, the predictive density $p(c|\varphi)$ is modeled using Laplace distributions:
\begin{equation}
\label{eq: laplace}
    \mathcal{L}(c|\sigma^2)=\frac{1}{\sqrt{2 \sigma^2}} e^{-\sqrt{\frac{2}{\sigma^2}}|c-\mu|} ,
\end{equation}
where the mean $\mu$ and variance $\sigma^2$ are predicted by the network as $(\mu,\sigma^2)=\Phi(\mathcal{I};\theta)$ for each point pair. $\sigma^2$ serves as the uncertainty mask, indicating higher confidence for matching pairs likely to overlap and lower confidence for non-overlapping or foreground regions. 

To predict the uncertainty mask, we first compute the cross-attention feature $\mathbf{G}^{\hat{\mathcal{P}}} \!\in\! \mathbb{R}^{\hat{N}\times{d_t}}$ for $\hat{\mathcal{P}}$ after obtaining the self-attention features $\mathbf{H}^{\hat{\mathcal{P}}}$ and $\mathbf{H}^{\hat{\mathcal{Q}}}$, \ie,
\begin{equation}
    \mathbf{G}_i^{\hat{\mathcal{P}}}\!=\!\sum\nolimits_{j=1}^{\hat{M}}a_{i,j}({\mathbf{H}_j^{\hat{\mathcal{Q}}}}\mathbf{W}^V),
\end{equation}
where $a_{i,j}\!=\!\operatorname{Softmax}(\frac{(\mathbf{H}_i^{\hat{\mathcal{P}}}\mathbf{W}^Q)(\mathbf{H}_j^{\hat{\mathcal{Q}}}\mathbf{W}^K)^T}{\sqrt{d_t}})$
is the weight coefficient computed by applying a row-wise softmax on the cross-attention score. Similarly, we can derive the cross-attention feature $\mathbf{G}^{\hat{\mathcal{Q}}}$ for $\mathcal{Q}$. 
$\mathbf{G}^{\hat{\mathcal{P}}}$ and $\mathbf{G}^{\hat{\mathcal{Q}}}$ exhibit robust geometric characteristics and awareness of hybrid motions between the input point clouds. 
Finally, we use an MLP, $\Phi$ with parameters $\theta$ followed by a sigmoid layer, to regress the uncertainty masks, \ie,
$\sigma^2_{\hat{\mathcal{P}}},\sigma^2_{\hat{\mathcal{Q}}}= \operatorname{Sigmoid}(\Phi(\mathbf{G}^{\hat{\mathcal{P}}}, \mathbf{G}^{\hat{\mathcal{Q}}};\theta)).$

\paragraph{Uncertainty-aware superpoint matching.}
To determine superpoint correspondences, we follow \cite{qin2022geometric} to first normalize $\mathbf{H}^{\hat{\mathcal{P}}}$ and $\mathbf{H}^{\hat{\mathcal{Q}}}$ onto a unit hypersphere.
Then, we compute the Gaussian correlation matrix $\mathbf{S} \!\in\! \mathbb{R}^{\hat{N}\times\hat{M}}$, where $s_{i,j}=\operatorname{exp}(-||\mathbf{h}_i^{\hat{\mathcal{P}}}-\mathbf{h}_i^{\hat{\mathcal{Q}}}||_2^2)$. 
Besides, a dual-normalization operation~\cite{rocco2018neighbourhood, sun2021loftr} is performed on $\mathbf{S}$ to suppress ambiguous matches.
To further enhance the robustness against incorrect matches caused by hybrid motions, we mask the correlation matrix $\mathbf{S}$ using our generated uncertainty masks. We then select the largest $N_c$ entries as the superpoint correspondences, \ie,
\begin{equation}
    \hat{\mathcal{C}}=\{(\hat{\mathbf{p}}_{x_i},\hat{\mathbf{q}}_{y_i})|(x_i,y_i)\in\operatorname{topk}_{x,y}(s_{x,y}\cdot (1-\sigma^2_{x,y}))\},
\end{equation}
where $\hat{\mathbf{p}}_{x_i}$ and $\hat{\mathbf{q}}_{y_i}$ are superpoints in $\hat{\mathcal{P}}$ and $\hat{\mathcal{Q}}$, respectively.

\paragraph{Transformation estimation.} 
To recover the 3D transformation between the inputs based on the estimated correspondences, we employ both the RANSAC algorithm and the deep robust estimator LGR proposed by~\cite{qin2022geometric}.

\begin{table*}[th]
    \centering
    \resizebox{\linewidth}{!}{
    \begin{tabular}{l|ccc|ccc|ccc|ccc|ccc|ccc}
        \toprule
         \multirow{4}{*}{Methods} & \multicolumn{9}{c|}{HybridMatch} & \multicolumn{9}{c}{HybridLoMatch} \\
         \cmidrule{2-19}
         & \multicolumn{3}{c|}{\textit{rigid-only}} & \multicolumn{3}{c|}{\textit{10-30\% non-rigid}} & \multicolumn{3}{c|}{\textit{30-50\% non-rigid}} & \multicolumn{3}{c|}{\textit{rigid-only}} & \multicolumn{3}{c|}{\textit{10-30\% non-rigid}} & \multicolumn{3}{c}{\textit{30-50\% non-rigid}}\\
         \cmidrule{2-19}
         & RRE $\downarrow$ & RTE $\downarrow$ & RR $\uparrow$ & RRE $\downarrow$ & RTE $\downarrow$ & RR $\uparrow$ & RRE $\downarrow$ & RTE $\downarrow$ & RR $\uparrow$ & RRE $\downarrow$ & RTE $\downarrow$ & RR $\uparrow$ & RRE $\downarrow$ & RTE $\downarrow$ & RR $\uparrow$ & RRE $\downarrow$ & RTE $\downarrow$ & RR $\uparrow$ \\
         \midrule
         PREDATOR$^\dagger$ & 1.7 & 6.7 & 98.7 & 4.6 & 19.2 & 31.6 & 5.6 & 24.5 & 26.6 & 2.0 & 7.6 & 89.9 & 9.7 & 34.9 & 6.2 & 8.2 & 30.4 & 5.0 \\
         CoFiNet$^\dagger$ & 1.3 & 5.2 & \underline{99.2} & 3.8 & 12.9 & 33.6 & 4.0 & 17.8 & 28.2 & 1.8 & 7.2 & 93.0 & 7.5 & 25.0 & 5.0 & 6.2 & 29.4 & 5.8 \\
         GeoTrans$^\dagger$ & 0.8 & 3.2 & 98.9 & 3.7 & 13.6 & 27.6 & 4.0 & 17.6 & 22.6 & \underline{1.1} & 4.2 & 91.3 & 5.7 & 18.3 & 5.2 & 5.7 & 25.5 & 1.4 \\
         \midrule
         PREDATOR$^\ddagger$ & 1.1 & 4.4 & 98.9 & 2.4 & 8.7 & 80.8 & 2.8 & 12.6 & 75.4 & 1.4 & 5.3 & \textbf{94.6} & 5.1 & 17.5 & 33.2 & 4.2 & 17.5 & 24.4 \\
         CoFiNet$^\ddagger$ & 1.4 & 5.8 & 98.2 & 2.8 & 10.6 & 82.4 & 3.3 & 14.6 & 78.0 & 1.8 & 7.3 & 88.9 & 5.1 & 18.0 & 29.8 & 4.7 & 23.1 & 25.6 \\
         GeoTrans$^\ddagger$ & \underline{0.6} & \underline{2.3} & \textbf{99.3} & \underline{1.4} & \underline{5.6} & \underline{95.6} & \underline{1.6} & \underline{8.1} & \underline{93.6} & \textbf{0.8} & \underline{3.4} & 93.8 & \underline{3.3} & \underline{12.3} & \underline{64.2} & \underline{3.2} & \underline{13.7} & \underline{61.0} \\
         Our \ourmethod{}$^\ddagger$ & \textbf{0.5} & \textbf{2.1} & \underline{99.2} & \textbf{1.2} & \textbf{5.4} & \textbf{97.4} & \textbf{1.5} & \textbf{7.1} & \textbf{96.6} & \textbf{0.8} & \textbf{3.3} & \underline{93.9} & \textbf{3.0} & \textbf{12.2} & \textbf{75.6} & \textbf{2.8} & \textbf{12.3} & \textbf{73.6} \\
         \bottomrule
    \end{tabular}}
    \caption{Evaluation results on our HybridMatch and HybridLoMatch.  $^{\dagger}$: the model is only trained on 3DMatch. $^{\ddagger}$: the model is trained on our HybridMatch. The best and second-best results are marked in bold and underlined for better comparison.}
    \label{tab: DeforMatch_LGR}
\end{table*}
\begin{table*}[h]
    \centering
    \resizebox{\linewidth}{!}{
    \begin{tabular}{l|ccccc|ccccc|ccccc|ccccc}
        \toprule
        \multirow{4}{*}{\# Samples} & \multicolumn{10}{c|}{HybridMatch} & \multicolumn{10}{c}{HybridLoMatch} \\
        \cmidrule{2-21}
         & \multicolumn{5}{c|}{\textit{10-30\% non-rigid}} & \multicolumn{5}{c|}{\textit{30-50\% non-rigid}} & \multicolumn{5}{c|}{\textit{10-30\% non-rigid}} & \multicolumn{5}{c}{\textit{30-50\% non-rigid}}\\
        \cmidrule{2-21}
        & 5000 & 2500 & 1000 & 500 & 250 & 5000 & 2500 & 1000 & 500 & 250 & 5000 & 2500 & 1000 & 500 & 250 & 5000 & 2500 & 1000 & 500 & 250 \\
        \midrule
        \multicolumn{21}{c}{Feature Matching Recall (\%) $\uparrow$} \\
        \midrule
        PREDATOR & 95.2 & 96.0 & 96.2 & 96.2 & 95.4 & 94.0 & 94.2 & 93.6 & 93.2 & 94.0 & 66.8 & \underline{68.4} & 68.4 & 70.2 & 67.4 & 62.8 & 64.0 & 64.8 & 63.8 & 61.8\\
        CoFiNet & 96.0 & \underline{96.4} & \underline{96.2} & \underline{96.4} & \underline{95.6} & 93.0 & 94.6 & \underline{95.2} & 95.2 & 95.0 & 59.4 & 61.4 & 63.2 & 63.4 & 64.0 & 53.4 & 55.2 & 58.8 & 59.4 & 59.4\\
        GeoTrans & \textbf{99.9} & \textbf{99.8} & \textbf{99.8} & \textbf{99.8} & \textbf{99.8} & \underline{99.4} & \underline{99.4} & \textbf{99.6} & \underline{99.4} & \underline{99.4} & \textbf{93.2} & \textbf{92.8} & \textbf{93.4} & \underline{92.2} & \underline{92.0} & \underline{91.4} & \underline{91.4} & \underline{90.4} & \underline{91.6} & \underline{90.8}\\
        \midrule
        Our \ourmethod{} & \underline{99.8} & \textbf{99.8} & \textbf{99.8} & \textbf{99.8} & \textbf{99.8} & \textbf{99.6} & \textbf{99.6} & \textbf{99.6} & \textbf{99.6} & \textbf{99.6} & \underline{92.8} & \textbf{92.8} & \underline{92.8} & \textbf{92.6} & \textbf{93.0} & \textbf{94.4} & \textbf{94.4} & \textbf{94.8} & \textbf{94.6} & \textbf{94.8}\\
        \midrule
        \multicolumn{21}{c}{Inlier Ratio (\%) $\uparrow$} \\
        \midrule
        PREDATOR & 32.4 & 34.8 & \underline{36.5} & \underline{36.7} & 35.1 & 29.4 & 31.6 & 33.1 & 32.9 & 31.8 & 11.4 & 12.2 & 13.0 & 13.2 & 12.6 & 9.8 & 10.5 & 11.1 & 11.0 & 10.9 \\
        CoFiNet & 23.4 & 26.7 & 31.2 & 32.9 & 34.0 & 21.3 & 24.5 & 28.7 & 30.5 & 31.6 & 7.7 & 8.7 & 10.9 & 11.8 & 12.4 & 7.1 & 7.9 & 9.7 & 10.6 & 11.2 \\
        GeoTrans & \underline{36.5} & \underline{36.5} & 36.4 & \underline{36.7} & \underline{36.4} & \underline{33.6} & \underline{33.6} & \underline{33.7} & \underline{33.6} & \underline{33.6} & \underline{18.2} & \underline{18.3} & \underline{18.3} & \underline{18.3} & \underline{18.4} & \underline{33.6} & \underline{17.3} & \underline{17.4} & \underline{17.3} & \underline{17.2} \\
        \midrule
        Our \ourmethod{} & \textbf{58.6} & \textbf{58.7} & \textbf{58.7} & \textbf{58.8} & \textbf{58.6} & \textbf{54.4} & \textbf{54.4} & \textbf{54.4} & \textbf{54.3} & \textbf{54.3} & \textbf{34.4} & \textbf{34.3} & \textbf{34.3} & \textbf{34.3} & \textbf{34.3} & \textbf{33.8} & \textbf{33.9} & \textbf{33.7} & \textbf{33.7} & \textbf{33.7}  \\
        \midrule
        \multicolumn{21}{c}{Registration Recall (\%) $\uparrow$} \\
        \midrule
        PREDATOR & 65.2 & 73.4 & 79.2 & 80.8 & 80.6 & 57.2 & 69.8 & 74.6 & 75.4 & 71.4 & 19.6 & 26.8 & 33.2 & 32.4 & 30.6& 15.6 & 21.2 & 24.2 & 22.8 & 24.4 \\
        CoFiNet & 79.0 & 80.4 & 82.4 & 80.6 & 79.8 & 75.6 & 78.0 & 77.8 & 77.2 & 72.2 & 27.2 & 27.8 & 29.8 & 27.6 & 27.2 & 21.6 & 24.8 & 25.2 & 25.6 & 24.8 \\
        GeoTrans & \underline{92.8} & \underline{91.4} & \underline{89.0} & \underline{88.6} & \underline{90.0} & \underline{91.0} & \underline{86.8} & \underline{88.6} & \underline{86.8} & \underline{86.4} & \underline{52.6} & \underline{51.4} & \underline{54.4} & \underline{49.2} & \underline{51.8} & \underline{52.4} & \underline{55.2} & \underline{43.8} & \underline{50.0} & \underline{50.8} \\
        \midrule
        Our \ourmethod{} & \textbf{97.4} & \textbf{96.4} & \textbf{97.6} & \textbf{97.2} & \textbf{96.4} & \textbf{96.6} & \textbf{97.4} & \textbf{97.8} & \textbf{95.0} & \textbf{96.0} & \textbf{75.6} & \textbf{73.0} & \textbf{75.2} & \textbf{73.0} & \textbf{70.8} & \textbf{76.2} & \textbf{75.8} & \textbf{75.0} & \textbf{74.2} & \textbf{72.4}  \\
        \bottomrule
    \end{tabular}}
    \caption{Evaluation results on HybridMatch and HybridLoMatch.
    }
    \label{tab: DeforMatch_RANSAC}
\end{table*}

\subsection{Loss Functions} 
\label{sec: 3.4}
The loss function consists of an uncertainty mask loss $\mathcal{L}_{um}$ for uncertainty prediction, an overlap-aware circle loss $\mathcal{L}_{oc}$ for superpoint matching, and a point matching loss $\mathcal{L}_{p}$.  
Both $\mathcal{L}_{oc}$ and $\mathcal{L}_{p}$ are implemented following~\cite{qin2022geometric}. 
The overall loss for training is $\mathcal{L}=\mathcal{L}_{um}+\mathcal{L}_{oc}+\mathcal{L}_{p}$.

\paragraph{Uncertainty mask loss $\mathcal{L}_{um}$.}
As customary in probabilistic regression~\cite{ilg2018uncertainty, kendall2017uncertainties, truong2023pdc}, we employ the negative log-likelihood loss to train the uncertainty mask generator:
\begin{equation}
    -\log p(\mathcal{C}|\Phi(\mathcal{I};\theta)\!=\!-\!\sum\nolimits_{i,j}\log p(c_{i,j}|\varphi_{i,j}(\mathcal{I};\theta)),
\end{equation}
where $p(c|\varphi)$ can be substituted by the Laplace distribution as detailed in Eq.~\ref{eq: laplace}. Taking into account both source and target uncertainty masks, $\mathcal{L}_{um}$ is formulated as
\begin{equation}
    \mathcal{L}_{um}\!=\!-\log(\sum\nolimits_{i,j}\!e^{\scriptscriptstyle{-\log(2)-\alpha_{i,j}-\sqrt{2}e^{-\frac{1}{2}\alpha_{i,j}}\cdot |c_{i,j}-\mu_{i,j}|}}),
\end{equation}
where $\alpha_{i,j} \!=\! \log(\sigma^2_{\hat{\mathcal{P}}} \!\cdot\! \sigma^2_{\hat{\mathcal{Q}}})_{i,j}$. 
Naturally, we use $\mu_{i,j} \!=\! s_{i,j}$ in the Gaussian correlation matrix $\mathbf{S}$ to denote the correspondence prediction for the point pair $i,j$.
To avoid division by zero, we use a numerically stable log-sum-exp function and introduce a regularization term $1-\sigma^2_{\hat{\mathcal{P}}} \!\cdot\! \sigma^2_{\hat{\mathcal{Q}}}$.

\paragraph{Overlap-aware circle loss $\mathcal{L}_{oc}$.}
To prioritize high-overlap matches, we compute the overlap-aware circle loss on $\mathcal{P}$,
\begin{equation}
    \mathcal{L}_{oc}^{\mathcal{P}} \! = \! \frac{1}{|\mathcal{A}|}  \!\!  \sum_{\mathcal{G}^{\mathcal{P}}_i \in \mathcal{A}} \!\!\! \log[1 + \! \! \! \sum_{\mathcal{G}^{\mathcal{Q}}_j \in \varepsilon_p^i} \!\!\!\!  e^{\lambda_i^j \beta_p^{i,j}(d_i^j - \Delta_p)} \cdot \!\!\!\! \sum_{\mathcal{P}^{\mathcal{Q}}_k\in \varepsilon_n^i} \!\!\!\! e^{\beta_p^{i,k}(\Delta_n - d_i^k)}].
\end{equation}
Here, $\mathcal{A}$ is the set of anchor patches in $\mathcal{P}$ that have at least one positive patch in $\mathcal{Q}$. 
For each anchor patch $\mathcal{G}^{\mathcal{P}}_i \! \in \! \mathcal{A}$, $\varepsilon_p^i$ and $\varepsilon_n^i$ denote the sets of its positive and negative patches in $\mathcal{Q}$. 
$d_i^j$ is the feature distance, and $\lambda_i^j = (o_i^j)^{\frac{1}{2}}$ and $o_i^j$ represents the overlap ratio between $\mathcal{G}^{\mathcal{P}}_i$ and $\mathcal{G}^{\mathcal{Q}}_j$. 
$\beta_p^{i,j} \!=\! \gamma(d_i^j \!-\! \Delta_p)$ and $\beta_n^{i,k} \!=\! \gamma(\Delta_n \!-\! d_i^k)$ are the positive and negative weights, respectively. 
The margins $\Delta_p$ and $\Delta_n$ are set to 0.1 and 0.4, respectively. The same goes for the loss $\mathcal{L}_{oc}^{\mathcal{Q}}$ on $\mathcal{Q}$.

\paragraph{Point matching loss $\mathcal{L}_{p}$}
is computed for the $i$th patch as
\begin{equation}
    \mathcal{L}_{p,i} = - \! \! \! \! \! \! \! \sum_{(u,v) \in \mathcal{M}_i} \! \! \! \! \! \! \log \Bar{z}_{\scriptscriptstyle{u,v}}^i - \! \! \! \sum_{u \in \mathcal{I}_i} \log \Bar{z}_{\scriptscriptstyle{u,m_i+\!1}}^i - \! \! \! \sum_{v \in \mathcal{J}_i} \log \Bar{z}_{\scriptscriptstyle{n_i+\!1, v}}^i,
\end{equation}
where $\mathcal{M}_i$ denotes the set of ground-truth point correspondences extracted with a matching radius $\tau$ from each $\hat{\mathcal{C}}_i^*$ and $\{ \hat{\mathcal{C}}_i^* \}$ denotes the randomly sampled $N_g$ ground-truth point correspondences. $\Bar{z}_{u,v}^i$ is the element in the $u$-th row and the $v$-th column of the soft assignment matrix $\Bar{\mathbf{Z}}_i$. $\mathcal{I}_i$ and $\mathcal{J}_i$ denote the sets of unmatched points in the two patches.

\section{Experiments}

\begin{figure*}[t]
    \centering
    \includegraphics[width=\linewidth]{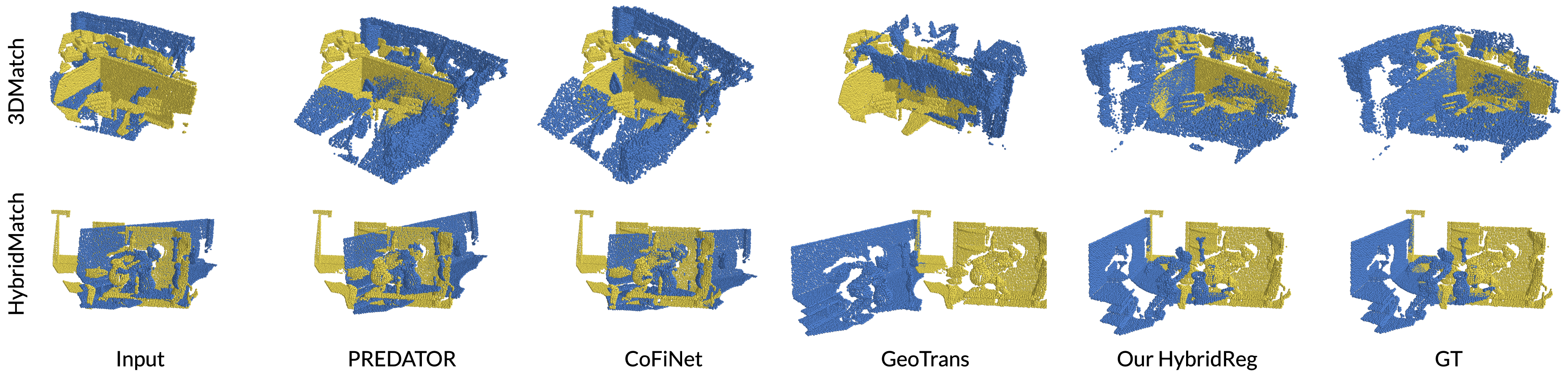}
    \caption{Qualitative comparison on \ourdataset{} and 3DMatch.}
    \label{fig: qualitative_DeforMatch_3DMatch}
\end{figure*}
\begin{table}[t]
    \centering
    \resizebox{\linewidth}{!}{
        \begin{tabular}{l|ccc|ccc}
        \toprule
        \multirow{2}{*}{Methods} & \multicolumn{3}{c|}{3DMatch} & \multicolumn{3}{c}{3DLoMatch} \\
        \cmidrule{2-7}
        & RRE $\downarrow$ & RTE $\downarrow$ & RR $\uparrow$ & RRE $\downarrow$ & RTE $\downarrow$ & RR $\uparrow$ \\
        \midrule
        PerfectMatch & 2.2 & 7.1 & 78.4 & 3.5 & 10.3 & 33.0 \\
        FCGF & \underline{2.0} & 6.6 & 85.1 & 3.2 & 10.0 & 40.1  \\
        D3Feat & 2.2 & 6.7 & 81.6 & 3.4 & 10.3 & 37.3 \\
        PREDATOR & \underline{2.0} & 6.4 & 89.0 & 3.1 & 9.3 & 59.8  \\
        CoFiNet & \underline{2.0} & 6.2 & 89.3 & 3.3 & 9.4 & 67.5 \\
        GeoTrans & \textbf{1.6} & 5.3 & 91.5 & \textbf{2.5} & \textbf{7.4} & \textbf{74.0} \\
        RegTR & \textbf{1.6} & \textbf{4.9} & \underline{92.0} & \underline{2.8} & 7.7 & 64.8 \\
        \midrule
        Our \ourmethod{} & \textbf{1.6} & \underline{5.2} & \textbf{92.7} & \textbf{2.5} & \underline{7.5} & \underline{73.1} \\
        \bottomrule
    \end{tabular}}
    \caption{Registration results on 3DMatch and 3DLoMatch.}
    \label{tab: 3DMatch}
\end{table}

\subsection{Datasets and Evaluation Metrics}
\paragraph{Datasets.}
To evaluate effectiveness on hybrid motions, we use the proposed \ourdataset{}/\ourlodataset{}, which is divided into three splits: \emph{rigid-only} (11.3k/14k pairs), \emph{10-30\% non-rigid} (7.3k/8.5k pairs), and \emph{30-50\% non-rigid} (4k/5.5k pairs). 
To assess robustness in indoor scenes, we employ the 3DMatch dataset~\cite{zeng20173dmatch}, comprising 62 scenes, with 46/8/8 scenes used for training/validation/testing. 3DMatch and 3DLoMatch are categorized based on $>$30\% and 10\%-30\% overlap ratios. 
To evaluate the generalizability in outdoor scenes, we transfer models trained on 3DMatch to the ETH dataset~\cite{pomerleau2012challenging}, including 4 scenes with 713 pairs of 132 point clouds.

\paragraph{Evaluation metrics.}
We adopt several metrics following~\cite{qin2022geometric,chen2023sira}: 
(i) Registration Recall (RR), the fraction of successfully registered pairs whose transformation error RMSE$<$0.2m/0.2m/0.5m for \ourdataset{}/3DMatch/ETH; 
(ii) Inlier Ratio (IR), the fraction of inlier correspondences with residuals $<$0.1m/0.1m/0.2m for \ourdataset{}/3DMatch/ETH; 
(iii) Feature Matching Recall (FMR), the fraction of pairs with IR$>$5\%; 
(iv) The median of the average Relative Rotation Error (RRE); and 
(v) The median of the average Relative Translation Error (RTE) for successfully registered pairs whose RMSE$<$0.2m/0.2m/0.5m for \ourdataset{}/3DMatch/ETH.

\subsection{Implementation Details}
To stabilize training, the uncertainty mask generator is first trained with L1 loss for 30 epochs, then fine-tuned with uncertainty mask loss for 20 epochs. 
Adam optimizer~\cite{kingma2014adam} is used with a batch size of 8 and a learning rate of 1e-4.
All experiments run on 8 NVIDIA Tesla P40 GPUs.
Please refer to our supp. material for more details.

\begin{figure}[t]
    \centering
    \includegraphics[width=\linewidth]{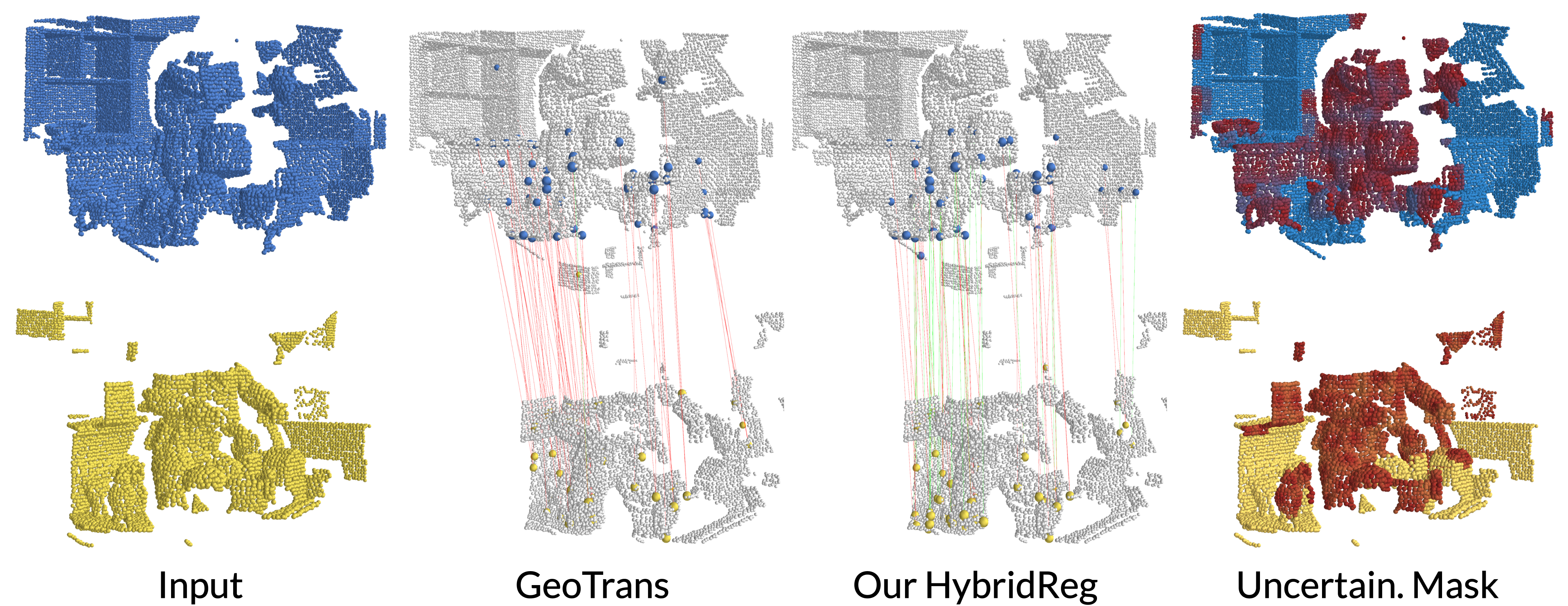}
    \caption{Visualization of correspondences and uncertainty masks, where unreliable non-rigid regions are marked in red.}
    \label{fig: qualitative_corr_mask}
\end{figure}

\subsection{Evaluation on \ourdataset{}}
We compare the quantitative registration results of our method with recent state-of-the-art approaches in Tab.~\ref{tab: DeforMatch_LGR} and Tab.~\ref{tab: DeforMatch_RANSAC}. 
As revealed in Tab.~\ref{tab: DeforMatch_LGR}, either transferring pre-trained model weights from 3DMatch to our \ourdataset{} \emph{rigid-only} split or direct training on our data yields high RR, validating our synthetic data.
However, models trained on rigid-only data struggle with hybrid motions, particularly in scenarios with significant foreground motion and low overlap.
Incorporating hybrid-motion data into training improves performance but remains inferior to rigid-only scenes, highlighting the need for algorithmic enhancements.
Our \ourmethod{} significantly outperforms other methods across almost all metrics, demonstrating robust registration capability for both rigid-only and challenging non-rigid scenarios. 
Specifically, our method achieves a 20.7\% increase in RR compared to the second best on the most challenging test split (\ourlodataset{}, \emph{30-50\% non-rigid}). 
Qualitative results in Fig.~\ref{fig: qualitative_DeforMatch_3DMatch} also illustrate our method's superior performance.
We visualize our uncertainty masks in Fig.~\ref{fig: qualitative_corr_mask}, where the precise masks improve the correspondence accuracy.

\begin{table*}[th]
    \centering
    \resizebox{\linewidth}{!}{
    \begin{tabular}{l|ccc|ccc|ccc|ccc}
        \toprule
         \multirow{4}{*}{Models} & \multicolumn{6}{c|}{HybridMatch} & \multicolumn{6}{c}{HybridLoMatch} \\
         \cmidrule{2-13}
         & \multicolumn{3}{c|}{\textit{10-30\% non-rigid}} & \multicolumn{3}{c|}{\textit{30-50\% non-rigid}} & \multicolumn{3}{c|}{\textit{10-30\% non-rigid}} & \multicolumn{3}{c}{\textit{30-50\% non-rigid}}\\
         \cmidrule{2-13}
         & RRE $\downarrow$ & RTE $\downarrow$ & RR $\uparrow$ & RRE $\downarrow$ & RTE $\downarrow$ & RR $\uparrow$ & RRE $\downarrow$ & RTE $\downarrow$ & RR $\uparrow$ & RRE $\downarrow$ & RTE $\downarrow$ & RR $\uparrow$ \\
         \midrule
         (a) Baseline (3D-FRONT) & 2.0 & 8.5 & 52.8 & 2.5 & 12.4 & 44.8 & 4.6 & 16.5 & 18.8 & 4.3 & 19.0 & 14.8 \\
         (b) + Non-rigid Foreground Motion & 1.6 & 7.5 & 90.4 & \underline{2.0} & 9.6 & 86.2 & 3.7 & 14.9 & 54.0 & 4.1 & 16.5 & 48.6 \\
         (c) + Rigid Foreground Motion & \underline{1.4} & \underline{6.0} & \underline{94.0} & \underline{2.0} & \underline{9.4} & \underline{89.2} & \underline{3.5} & \underline{14.1} & \underline{62.8} & \underline{3.3} & \underline{15.4} & \underline{54.2} \\
         (d) + Delete Planes &  \textbf{1.2} &  \textbf{5.4} &  \textbf{97.4} &  \textbf{1.5} &  \textbf{7.1} &  \textbf{96.6} &  \textbf{3.0} &  \textbf{12.2} &  \textbf{75.6} &  \textbf{2.8} &  \textbf{12.3} &  \textbf{73.6} \\
         \midrule
         (e) Baseline (GeoTrans) & \underline{1.4} & \underline{5.6} & 95.6 & \underline{1.6} & 8.1 & 93.6 & \underline{3.3} & \underline{12.3} & 64.2 & \underline{3.2} & 13.7 & 61.0 \\
         (f) w/ Uncertainty Mask + L1 Loss & \underline{1.4} & 5.9 & 94.2 & 1.7 & 8.6 & \underline{95.2} & 3.4 & 13.2 & 66.2 & 3.4 & 15.5 & 61.8 \\
         (g) w/ Uncertainty Mask + NLL Loss & \textbf{1.2} & \textbf{5.4} & \underline{97.2} & \textbf{1.5} & \underline{7.2} & \textbf{96.6} & \textbf{3.0} & \textbf{12.2} & \underline{75.5} & \textbf{2.8} & \underline{12.4} & \underline{73.2} \\
             (h) + Uncertainty-aware Correlation & \textbf{1.2} & \textbf{5.4} & \textbf{97.4} & \textbf{1.5} & \textbf{7.1} & \textbf{96.6} & \textbf{3.0} & \textbf{12.2} & \textbf{75.6} & \textbf{2.8} &  \textbf{12.3} & \textbf{73.6} \\
         \bottomrule
    \end{tabular}}
    \caption{Ablation studies of our proposed \ourdataset{} (top) and \ourmethod{} (bottom).}
    \label{tab: ablation_DeforMatch_HybridReg}
\end{table*}
\begin{table}[t]
    \centering
    \resizebox{\linewidth}{!}{
    \begin{tabular}{l|ccccc}
        \toprule
        \multicolumn{6}{c}{ETH} \\
        \midrule
        \# Samples & 5000 & 2500 & 1000 & 500 & 250  \\
        \midrule
        PerfectMatch & 81.4 & 73.5 & 59.3 & 46.5 & 35.0 \\
        D3Feat &59.1 & 50.4 & 49.7 & 44.6 & 29.1 \\
        FCGF &42.1 & 36.1 & 29.5 & 26.3 & 18.9 \\
        PREDATOR & 74.7 & 72.9 & 67.7 & 60.3 & 51.7 \\
        CoFiNet & 83.9 & 82.7 & 81.9 & 77.4 & 68.8 \\
        GeoTrans + RANSAC & 80.9 & 76.8 & 73.3 & 72.8 & 70.8 \\
        GeoTrans + LGR  & \multicolumn{2}{c}{\textemdash} & \underline{85.6} & \multicolumn{2}{c}{\textemdash} \\
        \midrule
        Our \ourmethod{} + LGR  & \multicolumn{2}{c}{\textemdash} & \textbf{86.5} & \multicolumn{2}{c}{\textemdash} \\
        \bottomrule
    \end{tabular}}
    \caption{Registration recall comparison on ETH. All models are trained on 3DMatch.
    \textemdash: results with different samples are not applicable for LGR since it uses all correspondences.
    }
    \label{tab: ETH}
\end{table}

\subsection{Evaluation on Rigid Benchmark}
\paragraph{Evaluation on indoor 3D(Lo)Match.}
Results on the rigid-only 3DMatch dataset are presented in Tab.~\ref{tab: 3DMatch}.
Our \ourmethod{} achieves state-of-the-art performance on 3DMatch and competitive results on 3DLoMatch, highlighting its ability to preserve the strength on rigid-only scenes effectively.
Fig.~\ref{fig: qualitative_DeforMatch_3DMatch} also indicates the effectiveness of our method. 

\begin{figure}[t]
    \centering
    \includegraphics[width=\linewidth]{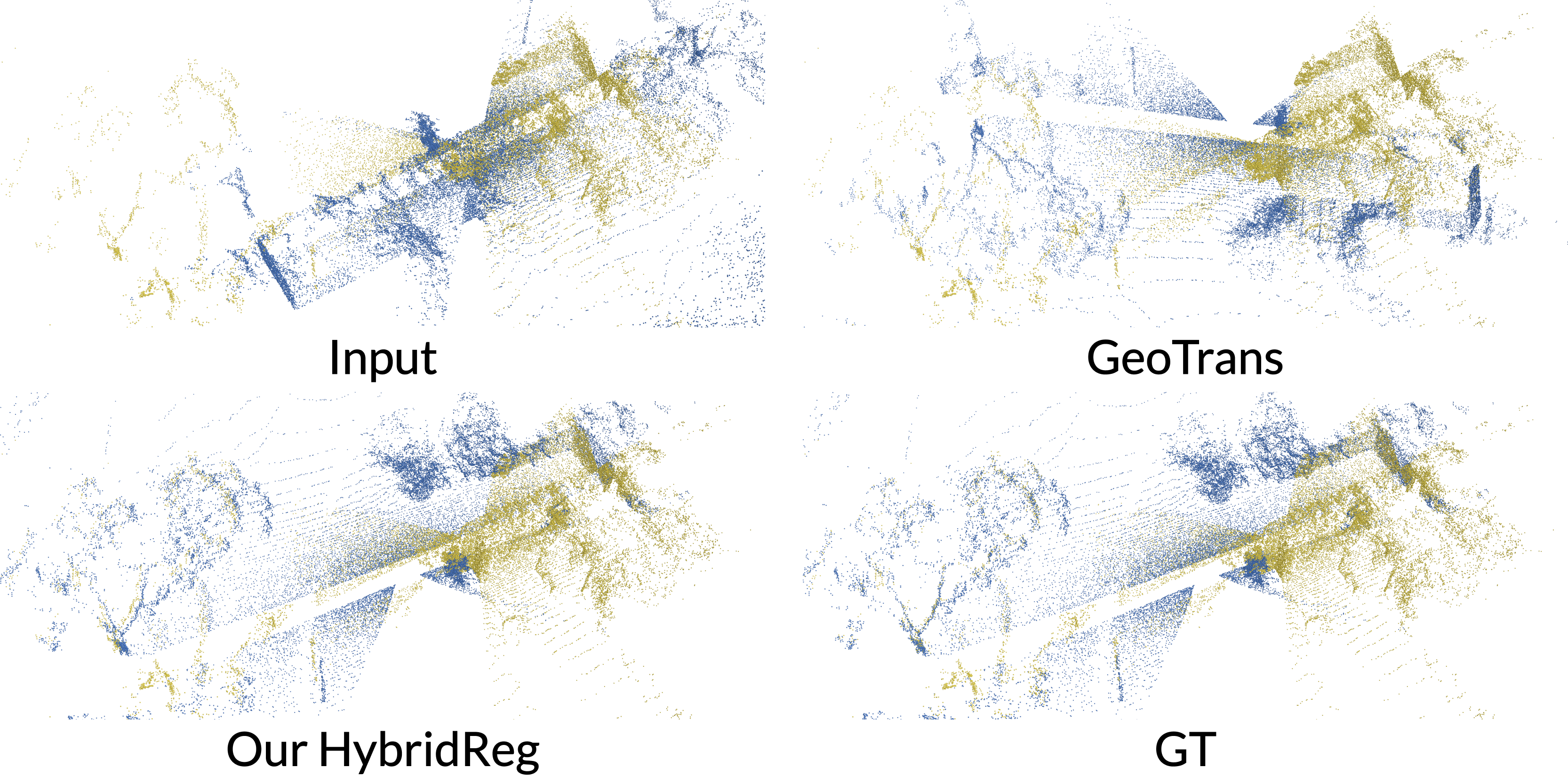}
    \caption{Qualitative comparison on ETH.}
    \label{fig: qualitative_ETH}
\end{figure}

\paragraph{Evaluation on outdoor ETH.}
To assess generalizability, we transfer models trained on 3DMatch to the ETH dataset, following~\cite{wang2023roreg, chen2023sira}. 
As shown in Tab.~\ref{tab: ETH}, our method outperforms the baseline (GeoTrans) and most other methods in RR, showcasing strong generalization capability in outdoor scenes. 
The qualitative comparison in Fig.~\ref{fig: qualitative_ETH} shows our method's effectiveness in addressing challenging cases with complex, unseen structures.

\subsection{Ablation Study}
We conduct extensive experiments on our \ourdataset{} and \ourmethod{} to evaluate the impact of each component. 
The hybrid-motion test set is fixed for consistent evaluation.

\paragraph{\ourdataset{}.}
We detail the construction of our \ourdataset{} in Tab.~\ref{tab: ablation_DeforMatch_HybridReg} (a-d).
By incrementally incorporating components into our baseline, we assess their contributions to the model's performance on the test set.
Comparing Rows (a) and (b), injecting non-rigid motions from DeformingThings4D significantly boosts registration performance.
Rows (c-d) and (b) demonstrate that adding rigid motions to selected foreground objects (c) and removing planes to balance geometric structures (d) help the model focus on background motion, thereby improving performance.

\paragraph{\ourmethod{}.}
To understand the efficacy of components in \ourmethod{}, we ablate each of them as shown in Tab.~\ref{tab: ablation_DeforMatch_HybridReg} (e-h).
Initially, we integrate the uncertainty mask learning module into the baseline, employing two loss variants for supervision.
Comparing Rows (f-g) and (e), the negative log-likelihood (NLL) loss outperforms the L1 loss in enhancing uncertainty modeling and learning capabilities, leading the model to better identify inlier correspondences, such as rigid background and immovable foreground. 
Further, comparing Rows (g) and (h), using the predicted uncertainty mask to refine the Gaussian correlation matrix yields an additional performance boost by suppressing incorrect matches.

\section{Conclusion}
We introduce \ourmethod{}, a novel 3D point cloud registration framework to account for hybrid motions and to enhance robustness.
To our best knowledge, we create the first large-scale indoor synthetic dataset, \ourdataset{}, tailored for point cloud registration with hybrid motions.
In addition, we propose a probabilistic uncertainty mask learning module to mitigate the impact of outliers, guided by an effective negative log-likelihood loss, enhancing feature extraction and correspondence matching.
Experimental quantitative and qualitative results manifest the state-of-the-art performance of \ourmethod{} consistently on various indoor and outdoor benchmarks.
%

\section*{Acknowledgments}
This work was supported in part by the National Natural Science Foundation of China (NSFC) under Grant Nos. 62372091, 62071097, 62236007, the Sichuan Science and Technology Program under Grant Nos. 2023NSFSC0462, 2023NSFSC0458, 2023NSFSC1972, and the Research Grants Council of the Hong Kong Special Administrative Region, China (Project No. T45-401/22-N).

\bibliography{aaai25}
\end{document}


\maketitle

\input{sections/supp}

%

\bibliography{aaai25}

%% file: sections/supp.tex
\section{More Qualitative Results}
Please refer to our video for more qualitative results and visualizations. 
Note that the video quality is reduced due to the file size limitation of the supplementary material.

\section{More Experimental Results}
This section presents additional experiments that evaluate our method and more experimental details.
\paragraph{Comparison on \ourdataset{}.}
We compare with another three recent methods, D3Feat~\cite{bai2020d3feat}, BUFFER~\cite{Ao_2023_CVPR}, and Lepard~\cite{lepard2021} on our \ourdataset{}, as shown in Tab.~\ref{tab: DeforMatch_latest_model}.
For fairness, all methods are trained on our \ourdataset{} using their official-released codes without modification.
We can see that they all perform well in rigid-only scenes, however, due to the lack of designs for hybrid motions, their performances degrade heavily with increasing non-rigid proportion.
Our \ourmethod{} shows robustness across different types of motions.

\paragraph{Comparison on 3DMatch.}
Following~\cite{huang2020predator, qin2022geometric}, we evaluate scene-wise registration on the 3DMatch and 3DLoMatch datasets. Tab.~\ref{table: scene-wise} reports the results. 
Since extremely large errors generated in failure cases can easily dominate the results, median RRE and RTE for each scene are reported only on the successfully-registered pairs. 
As RegTR~\cite{yew2022regtr} did not report such scene-wise results, we present the reproduced results using the released model from~\cite{chen2023sira}.
Our method achieves {\em the best performance\/} (see the ``mean'' columns) for {\em almost all cases\/} (two datasets and three metrics).

\paragraph{Comparison on ETH.}
We report complete quantitative results on ETH in Tab.~\ref{tab: ETH_full}.
FMR and IR only reflect the correspondence quality vs. RR directly reflects the registration accuracy. 
Similar to GeoTrans~\cite{qin2022geometric}, our method achieves high RR with low FMR \& IR, as LGR is relatively insensitive to outliers compared with RANSAC.
LGR uses the local-to-global strategy, so the transformation can be accurately estimated only from precisely-matched correspondences in a small set of local patches, leading to higher RR, even with lower FMR \& IR.

\paragraph{Ablation study.}
In Tab.~\ref{tab: ablation_RANSAC}, we present supplementary ablation results on our \ourdataset{} using RANSAC as the transformation estimator. 
In Tab.~\ref{tab: ablation_superpoint_topk}, we show ablation results on the number of correspondences in the Top-k selection.
\input{tables/DeforMatch_LGR_lateast_model}
\input{tables/3DMatch_split_scene}

\input{tables/ETH_full}
\input{tables/ablation_superpoint_topk}
\input{tables/ablation_RANSAC}

\section{Implementation details.}
\paragraph{Datas source of \ourdataset{}.}
We construct indoor scene backgrounds using 3D-FRONT, a large-scale dataset of synthetic 3D indoor scenarios, containing 18,968 rooms with 13,151 CAD 3D furniture objects. 
The layouts of the rooms are created by professional designers and distinctively span 31 scene categories and 34 object semantic super-classes. 
To enrich the geometric structure, two other datasets are applied as complementary components. 
ShapeNet is a richly-annotated large-scale repository of object-level 3D CAD models. 
It contains 55 common object categories, including about 51,300 unique 3D models. DeformingThings4D is an synthetic dataset, containing 1,972 animation sequences spanning 31 categories of humanoids and animals.

\paragraph{Rendering details.}
For each scene in our \ourdataset{} dataset, we randomly select 36 human animation sequences and 96 animal animation sequences from the DeformingThings4D dataset.
We use Blender to construct a 3D-FRONT scene, where objects from ShapeNet are randomly placed on 
top of randomly selected furniture, such as tables, chairs, or beds. Each object is dropped from a height of one meter above the furniture and locked into place once it comes to rest. 
We place one to seven 
random objects per furniture piece.

To obtain diverse and rich camera viewpoints, each piece of furniture is assigned a score and the viewpoint score is calculated by summing the scores of all visible furniture from that viewpoint. 
We set the threshold of the score to 0.8 and repeat 10,000 times to obtain 12 optimal camera poses. 
To ensure sufficient space for placing deforming animals and humans, we sample camera poses at only 1.2 meters away from furniture or walls. 
The camera resolution is set to $640\!\times\!480$, with a render sample count of 128.

Next, we randomly select one or two human animation sequences from the 36 chosen, and zero or one animal animation sequence from the 96 chosen. 
To increase the variety in non-rigid deformation, each frame is assigned a random X-Y position and Z rotation offset. 
Humans and animals are placed two meters away from the camera, with a minimum distance of 0.5 meters between them. 
One of the human animations is designated as the key animation, which the camera follows frame by frame. 
To augment the scene, 10-20 objects from ShapeNet are randomly placed around the key animation within a 3-meter radius, with height limits ranging from 0.1 to 2.5 meters. 
The positions and rotations of these objects are adjusted frame by frame to match the key animation. 
Additionally, movable furniture is adjusted within the current viewpoint, using similar offset sequences as the flying objects.

We calculate the deformation change coefficient for the animation by taking the absolute value of the offset for each vertex and averaging them. To render the data for registration, we track the non-rigid transformation of the key animation. We only select frames where the change coefficient compared to the previously selected frame is greater than 0.25 meters, excluding random position and rotation offsets, to ensure meaningful movement. 
For each frame, the camera is given a random location offset and is calibrated to focus on the key animation. 
The vertex offsets provided by DeformingThings4D are applied to the human and animal models to achieve deformation, followed by random position and rotation offsets. 
The flying objects and movable furniture are also updated to reflect the new frame. 
We use two paired frames to get a source and target point cloud pair. 
We calculate the animation proportion by comparing the depth differences between the current frame and a frame with all moving targets made invisible. 
We then calculate the background overlap ratio to closely approximate the real-world conditions. 
Finally, the rigid-only pairs are generated by moving only the camera while freezing deformable and movable subjects.

This integrated process ensures that our \ourdataset{} dataset effectively captures the complexity of real-world indoor environments with a mix of rigid and non-rigid motions, providing diverse and realistic scenarios for analysis.

\paragraph{Evaluation metrics.}
Following the existing works~\cite{qin2022geometric,chen2023sira}, we present the definitions of the evaluation metrics we employed.

\noindent 
(i) {\bf Registration Recall (RR)\/} is the fraction of successfully-registered point cloud pairs. A point cloud pair is said to be successfully registered when its transformation error is lower than threshold $\tau_1$. In addition, the transformation error is defined as the root mean square error of the ground-truth correspondences $\mathcal{C}^*$, to which the estimated transformation $\mathbf{T}_{est}(\cdot)$ has applied:
\begin{equation}
    \mathrm{RMSE} = \sqrt{\frac{1}{|\mathcal{C}^*|} \sum_{(\mathbf{p}_{x}^*, \mathbf{q}_{y}^*) \in \mathcal{C}^*} \| \mathbf{T}_{est}(\mathbf{p}_{x}^*) - \mathbf{q}_{y}^*\|_2^2 },
\end{equation}
\begin{equation}
    \mathrm{RR} = \frac{1}{M}\sum_{i=1}^M  \big [ \mathrm{RMSE}_i < \tau_1 \big ],
\end{equation}
where $\mathbf{p}_{x}$ and $\mathbf{q}_{y}$ denote the $x$-th point in source $\mathcal{P}$ and $y$-th point in target $\mathcal{Q}$, respectively; $[\cdot]$ is the Iversion bracket; and $M$ is the number of all point cloud pairs.

\noindent 
(ii) {\bf Inlier Ratio (IR)\/} is the fraction of inlier correspondences among all hypothesized correspondences $\mathcal{C}$. A correspondence is regarded as an inlier, if the distance between the two points is lower than a certain threshold $\tau_2$ under the ground-truth transformation $\mathbf{T}_{gt}(\cdot)$:
\begin{equation}
    \mathrm{IR} = \frac{1}{|\mathcal{C}|} \sum_{(\mathbf{p}_{x}, \mathbf{q}_{y}) \in \mathcal{C}} \big [ \|  \mathbf{T}_{gt}(\mathbf{p}_{x}) - \mathbf{q}_{y} \|_2 < \tau_2 \big ].
\end{equation}

\noindent 
(iii) {\bf Feature Matching Recall (FMR)\/} is the fraction of point cloud pairs whose IR $>$ threshold $\tau_3$:
\begin{equation}
    \mathrm{FMR} = \frac{1}{M}\sum_{i=1}^{M} \big [\mathrm{IR}_i > \tau_3 \big ].
\end{equation}

\noindent 
(iv) {\bf Relative Rotation Error (RRE)\/} is the geodesic distance in degrees between the estimated and ground-truth rotation matrices $\mathbf{R}_{est}$ and $\mathbf{R}_{gt}$:
\begin{equation}
    \mathrm{RRE} = \arccos \left ( \frac{\mathrm{trace}(\mathbf{R}_{est}^T \cdot \mathbf{R}_{gt} - 1)}{2} \right ).
\end{equation}

\noindent 
(v) {\bf Relative Translation Error (RTE)\/} is the Euclidean distance between estimated and ground-truth translation vectors $\mathbf{t}_{est}$ and $\mathbf{t}_{gt}$:
\begin{equation}
    \mathrm{RTE} = \| \mathbf{t}_{est} - \mathbf{t}_{gt} \|_2.
\end{equation}

Following~\cite{bai2020d3feat, huang2020predator, qin2022geometric, yew2022regtr}, we set $\tau_1=0.2\mathrm{m}$, $\tau_2=0.1\mathrm{m}$, and $\tau_3=0.05$ for evaluation on \ourdataset{}, \ourlodataset{}, 3DMatch, and 3DLoMatch benchmarks. 
Following~\cite{wang2023roreg}, we set $\tau_1=0.5\mathrm{m}$, $\tau_2=0.2\mathrm{m}$, and $\tau_3=0.05$ for evaluation on the ETH benchmark.

%% file: tables/DeforMatch_LGR_lateast_model.tex
\begin{table*}[t]
    \centering
    \setlength{\tabcolsep}{2pt}
    \resizebox{\linewidth}{!}{
    \begin{tabular}{l|ccc|ccc|ccc|ccc|ccc|ccc}
        \toprule
         \multirow{4}{*}{Methods} & \multicolumn{9}{c|}{HybridMatch} & \multicolumn{9}{c}{HybridLoMatch} \\
         \cmidrule{2-19}
         & \multicolumn{3}{c|}{\textit{rigid-only}} & \multicolumn{3}{c|}{\textit{10-30\% non-rigid}} & \multicolumn{3}{c|}{\textit{30-50\% non-rigid}} & \multicolumn{3}{c|}{\textit{rigid-only}} & \multicolumn{3}{c|}{\textit{10-30\% non-rigid}} & \multicolumn{3}{c}{\textit{30-50\% non-rigid}}\\
         \cmidrule{2-19}
         & RRE $\downarrow$ & RTE $\downarrow$ & RR $\uparrow$ & RRE $\downarrow$ & RTE $\downarrow$ & RR $\uparrow$ & RRE $\downarrow$ & RTE $\downarrow$ & RR $\uparrow$ & RRE $\downarrow$ & RTE $\downarrow$ & RR $\uparrow$ & RRE $\downarrow$ & RTE $\downarrow$ & RR $\uparrow$ & RRE $\downarrow$ & RTE $\downarrow$ & RR $\uparrow$ \\
         \midrule
         D3Feat &1.2 &5.0 &91.8 & \underline{3.0} & 19.0 & 20.1 & \underline{3.5} & 22.2 & 24.4 &1.8 &7.1 & 70.5 & \underline{3.8} & \underline{19.4} & 3.4 & \underline{3.7} & 21.7 & \underline{5.4} \\
         BUFFER &3.9 &13.1 &87.4 & 5.2 & 24.1 & \underline{37.4} & 6.2 & 26.4 & \underline{34.2} &4.3 &18.0 &53.9 & 6.4 & 29.9 & \underline{6.6} & 5.9 & \underline{20.3} & \underline{5.4} \\
         Lepard &\underline{1.1} &\underline{4.4} &\underline{96.6} & 4.6 & \underline{18.8} & 19.6 & 4.4 & \underline{15.8} & 20.2 &\underline{1.4} &\underline{5.8} &\underline{78.1} & 5.2 & 26.8 & 3.6 & 5.1 & 26.4 & 3.2 \\
        \midrule
         Our \ourmethod{} & \textbf{0.5} & \textbf{2.1} & \textbf{99.2} & \textbf{1.2} & \textbf{5.4} & \textbf{97.4} & \textbf{1.5} & \textbf{7.1} & \textbf{96.6} & \textbf{0.8} & \textbf{3.3} & \textbf{93.9} & \textbf{3.0} & \textbf{12.2} & \textbf{75.6} & \textbf{2.8} & \textbf{12.3} & \textbf{73.6} \\
         \bottomrule
    \end{tabular}}
    \caption{Evaluation results on our HybridMatch and HybridLoMatch.  All models are trained on our HybridMatch. }
    \label{tab: DeforMatch_latest_model}
\end{table*}

%% file: tables/3DMatch_split_scene.tex
\begin{table*}[!t]
    \scriptsize
    \setlength{\tabcolsep}{0.5pt}
    \centering
    \resizebox{\linewidth}{!}{
    \begin{tabular}{l|ccccccccc|ccccccccc}
        \toprule
        \multirow{2}{*}{Model} & \multicolumn{9}{c|}{3DMatch} & \multicolumn{9}{c}{3DLoMatch} \\
        \cmidrule{2-19}
         & Kitchen & Home1 & Home2 & Hotel1 & Hotel2 & Hotel3 & Study & Lab & Mean & Kitchen & Home1 & Home2 & Hotel1 & Hotel2 & Hotel3 & Study & Lab & Mean \\
        \midrule
        \multicolumn{19}{c}{\emph{Registration Recall} (\%) $\uparrow$} \\
        \midrule
        PerfectMatch & 90.6 & 90.6 & 65.4 & 89.6 & 82.1 & 80.8 & 68.4 & 60.0 & 78.4 & 51.4 & 25.9 & 44.1 & 41.1 & 30.7 & 36.6 & 14.0 & 20.3 & 33.0 \\
        FCGF & 98.0 & 94.3 & 68.6 & 96.7 & 91.0 & 84.6 & 76.1 & 71.1 & 85.1 & 60.8 & 42.2 & 53.6 & 53.1 & 38.0 & 26.8 & 16.1 & 30.4 & 40.1 \\
        D3Feat & 96.0 & 86.8 & 67.3 & 90.7 & 88.5 & 80.8 & 78.2 & 64.4 & 81.6 & 49.7 & 37.2 & 47.3 & 47.8 & 36.5 & 31.7 & 15.7 & 31.9 & 37.2 \\
        PREDATOR & 97.6 & 97.2 & 74.8 & \textbf{98.9} & \textbf{96.2} & 88.5 & 85.9 & 73.3 & 89.0 & 71.5 & 58.2 & 60.8 & 77.5 & 64.2 & 61.0 & 45.8 & 39.1 & 59.8 \\
        CoFiNet & 96.4 & \underline{99.1} & 73.6 & 95.6 & 91.0 & 84.6 & \underline{89.7} & 84.4 & 89.3 & 76.7 & 66.7 & 64.0 & 81.3 & 65.0 & 63.4 & 53.4 & 69.6 & 67.5 \\
        GeoTrans & \textbf{98.9} & 97.2 & \underline{81.1} & \textbf{98.9} & 89.7 & 88.5 & 88.9 & \underline{88.9} & 91.5 & \textbf{85.9} & \textbf{73.5} & \underline{72.5} & \textbf{89.5} & \underline{73.2} & \textbf{66.7} & \textbf{55.3} & \underline{75.7} & \textbf{74.0} \\
        RegTR$^*$ & 97.8 & 90.6 & 75.5 & 97.8 & 94.9 & \textbf{100.0} & 88.5 & \textbf{91.1} & \underline{92.0} & 66.0 & 58.2 & 64.9 & 72.7 & 61.3 & \textbf{70.7} & 53.4 & 71.0 & 64.8 \\
        \midrule
        Our HybridReg & \underline{98.3} & \textbf{99.2} & \textbf{81.7} & \underline{98.8} & \underline{95.3} & \underline{92.3} & \textbf{90.6} & 85.4 & \textbf{92.7} & \underline{84.5} & \underline{72.8} & \textbf{73.3} & \underline{85.5} & \textbf{74.2} & 64.7 & \underline{54.0} & \textbf{75.8} & \underline{73.1} \\
        \midrule
        \multicolumn{19}{c}{\emph{Relative Rotation Error} ($^{\circ}$) $\downarrow$} \\
        \midrule
        PerfectMatch & 1.9 & 1.8 & 2.3 & 2.0 & 2.0 & 2.9 & 2.3 & 2.3 & 2.2 & 3.0 & 3.9 & 3.4 & 3.2 & 3.2 & 3.3 & 4.3 & 3.8 & 3.5 \\
        FCGF & \underline{1.8} & 1.8 & \underline{2.2} & 1.9 & 1.7 & 2.4 & \underline{2.0} & 1.8 & \underline{1.9} & \underline{2.9} & 3.2 & 3.3 & 2.8 & 2.8 & \underline{2.8} & 3.4 & 4.0 & 3.1 \\
        D3Feat & 2.0 & 2.0 & 2.4 & 2.0 & 2.0 & 2.4 & 2.3 & 2.1 & 2.2 & 3.2 & 3.5 & 3.4 & 3.3 & 3.2 & 3.0 & 3.7 & 3.6 & 3.4 \\
        PREDATOR & 1.9 & 1.8 & 2.5 & 2.0 & \underline{1.6} & 2.5 & 2.1 & 1.9 & 2.0 & 3.1 & 2.6 & 3.2 & \underline{2.7} & 2.9 & 3.4 & \underline{3.0} & 3.4 & 3.0 \\
        CoFiNet & 1.9 & 1.8 & 2.3 & 1.8 & 1.8 & \textbf{1.6} & 2.5 & 2.3 & 2.0 & 3.2 & 3.1 & 3.7 & 2.8 & 2.9 & 3.2 & 4.1 & 3.1 & 3.3 \\
        GeoTrans & 1.8 & \underline{1.4} & \textbf{1.8} & \underline{1.5} & \textbf{1.3} & \textbf{1.6} & \underline{2.0} & 1.7 & \textbf{1.6} & \textbf{2.4} & \textbf{2.3} & \textbf{2.5} & \textbf{2.5} & \underline{2.5} & \textbf{2.5} & \underline{3.0} & 2.7 & \textbf{2.5} \\
        RegTR$^*$ & \textbf{1.7} & \underline{1.4} & \textbf{1.8} & 1.7 & \textbf{1.3} & \underline{1.8} & \textbf{1.6} & \textbf{1.4} & \textbf{1.6} & 3.4 & \underline{2.5} & 3.2 & \underline{2.7} & \textbf{2.4} & 3.0 & 3.1 & \underline{2.4} & \underline{2.8} \\
        \midrule
        Our HybridReg & \underline{1.8} & \textbf{1.3} & \textbf{1.8} & \textbf{1.4} & \textbf{1.3} & \textbf{1.6} & \underline{2.0} & \underline{1.6} & \textbf{1.6} & \textbf{2.4} & \textbf{2.3} & \underline{2.7} & \textbf{2.5} & \underline{2.5} & \textbf{2.5} & \textbf{2.8} & \textbf{2.3} & \textbf{2.5} \\
        
        \midrule
        \multicolumn{19}{c}{\emph{Relative Translation Error} (cm) $\downarrow$} \\
        \midrule
        PerfectMatch & 5.9 & 7.0 & 7.9 & 6.5 & 7.4 & 6.2 & 9.3 & 6.5 & 7.1 & 8.2 & 9.8 & 9.6 & 10.1 & 8.0 & 8.9 & 15.8 & 12.0 & 10.3 \\
        FCGF & 5.3 & 5.6 & 7.1 & 6.2 & 6.1 & 5.5 & 8.2 & 9.0 & 6.6 & 8.4 & 9.7 & 7.6 & 10.1 & 8.4 & 7.7 & 14.4 & 14.0 & 10.0 \\
        D3Feat & 5.5 & 6.5 & 8.0 & 6.4 & 7.8 & 4.9 & 8.3 & 6.4 & 6.7 & 8.8 & 10.1 & 8.6 & 9.9 & 9.2 & 7.5 & 14.6 & 13.5 & 10.3 \\
        PREDATOR & 4.8 & 5.5 & 7.0 & 7.3 & 6.0 & 6.5 & 8.0 & 6.3 & 6.4 & 8.1 & 8.0 & 8.4 & 9.9 & 9.6 & 7.7 & \underline{10.1} & 13.0 & 9.3 \\
        CoFiNet & 4.7 & 5.9 & 6.3 & 6.3 & 5.8 & \underline{4.4} & 8.7 & 7.5 & 6.2 & 8.0 & 7.8 & 7.8 & 9.9 & 8.6 & 7.7 & 13.1 & 12.3 & 9.4 \\
        GeoTrans & 4.2 & \underline{4.6} & 5.9 & \underline{5.5} & \underline{4.6} & 5.0 & 7.3 & \textbf{5.3} & 5.3 & \textbf{6.2} & 7.0 & \textbf{7.1} & \underline{8.0} & \underline{7.5} & \textbf{4.9} & 10.7 & 8.3 & \textbf{7.4} \\
        RegTR$^*$ & \textbf{4.0} & \textbf{4.1} & \underline{5.8} & 5.7 & \textbf{4.2} & \textbf{3.9} & \textbf{5.3} & \underline{5.7} & \textbf{4.8} & 8.0 & \textbf{6.4} & 7.7 & 9.3 & \textbf{7.3} & 6.0 & \textbf{9.4} & \underline{7.9} & 7.7 \\
        \midrule
        Our HybridReg & \underline{4.1} & \underline{4.6} & \textbf{5.1} & \textbf{5.0} & 4.9 & 4.5 & \underline{7.1} & 5.8 & \underline{5.2} & \underline{6.6} & \underline{7.0} & \underline{7.4} & \textbf{7.9} & \textbf{7.3} & \underline{5.9} & \underline{10.1} & \textbf{7.6} & \underline{7.5} \\
        \bottomrule

    \end{tabular}}
    \caption{Scene-wise registration results on the 3DMatch and the 3DLoMatch. $^*$: the results produced using the released model. For better comparison, the best and second-best results are marked in \textbf{bold} and \underline{underlined}.}
    \label{table: scene-wise}
\end{table*}

%% file: tables/ETH_full.tex
\begin{table*}[th]
\setlength{\tabcolsep}{6.5pt}
    \centering
    \resizebox{\linewidth}{!}{
    \begin{tabular}{l|ccccc|ccccc|ccccc}
        \toprule
        \multicolumn{16}{c}{ETH} \\
        \midrule
        \multirow{2}{*}{\# Samples} & \multicolumn{5}{c|}{Feature Matching Recall (\%) $\uparrow$} & \multicolumn{5}{c|}{Inlier Ratio (\%) $\uparrow$} & \multicolumn{5}{c}{Registration Recall (\%) $\uparrow$} \\
        \cmidrule{2-16}
        & 5000 & 2500 & 1000 & 500 & 250  & 5000 & 2500 & 1000 & 500 & 250  & 5000 & 2500 & 1000 & 500 & 250 \\
        \midrule
        PerfectMatch & \textbf{95.6} & \textbf{94.3} & \underline{80.5} & 69.1 & 51.4 & \textbf{19.7} &\textbf{ 16.7} & \underline{12.4} & 9.3 & 6.6 & 81.4 & 73.5 & 59.3 & 46.5 & 35.0 \\
        D3Feat & 63.3 & 71.0 & 69.7 & 67.9 & 60.5 & \underline{12.5} & \underline{13.2} & \textbf{13.6} & \underline{13.5} &\underline{ 11.8} & 59.1 & 50.4 & 49.7 & 44.6 & 29.1 \\
        FCGF & 41.1 & 38.4 & 32.3 & 24.6 & 15.9 & 5.8 & 5.3 & 4.4 & 3.5 & 2.8 & 42.1 & 36.1 & 29.5 & 26.3 & 18.9 \\
        PREDATOR & 65.6 & 64.5 & 59.6 & 52.0 & 40.5 & 11.1 & 10.3 & 8.5 & 6.8 & 5.1 & 74.7 & 72.9 & 67.7 & 60.3 & 51.7 \\
        CoFiNet & \underline{82.5} & \underline{83.7} & \textbf{81.9} & \textbf{81.1} & \textbf{79.9} & 9.6 & 9.8 & 9.9 & 9.9 & 9.8 & 83.9 & 82.7 & 81.9 & 77.4 & 68.8 \\
        GeoTrans + RANSAC & 60.5 & 67.5 & 75.7 & \underline{79.0} & \underline{79.8} & 6.9 & 9.2 & 12.1 & \textbf{13.7} & \textbf{14.6} & 80.9 & 76.8 & 73.3 & 72.8 & 70.8 \\
        GeoTrans + LGR & \multicolumn{2}{c}{\textemdash} & 59.9 & \multicolumn{2}{c|}{\textemdash} & \multicolumn{2}{c}{\textemdash} & 6.7 & \multicolumn{2}{c|}{\textemdash} & \multicolumn{2}{c}{\textemdash} & \underline{85.6} & \multicolumn{2}{c}{\textemdash} \\
        \midrule
        Our \ourmethod{} + LGR & \multicolumn{2}{c}{\textemdash} & 63.4 & \multicolumn{2}{c|}{\textemdash} & \multicolumn{2}{c}{\textemdash} & 7.0 & \multicolumn{2}{c|}{\textemdash} & \multicolumn{2}{c}{\textemdash} & \textbf{86.5} & \multicolumn{2}{c}{\textemdash} \\
        \bottomrule
    \end{tabular}}
    \caption{Evaluation results on ETH. All models are trained on 3DMatch. 
    \textemdash: results with different samples are not applicable for LGR since it uses all correspondences.
    }
    \label{tab: ETH_full}
\end{table*}

%% file: tables/ablation_superpoint_topk.tex
\begin{table*}[th]
    \centering
    \setlength{\tabcolsep}{10pt}
    \resizebox{\linewidth}{!}{
    \begin{tabular}{c|ccc|ccc|ccc|ccc}
        \toprule
         \multirow{4}{*}{\#Top-K Corr.} & \multicolumn{6}{c|}{\ourdataset{}} & \multicolumn{6}{c}{\ourlodataset{}} \\
         \cmidrule{2-13}
         & \multicolumn{3}{c|}{\textit{10-30\% non-rigid}} & \multicolumn{3}{c|}{\textit{30-50\% non-rigid}} & \multicolumn{3}{c|}{\textit{10-30\% non-rigid}} & \multicolumn{3}{c}{\textit{30-50\% non-rigid}}\\
         \cmidrule{2-13}
         & RRE $\downarrow$ & RTE $\downarrow$ & RR $\uparrow$ & RRE $\downarrow$ & RTE $\downarrow$ & RR $\uparrow$ & RRE $\downarrow$ & RTE $\downarrow$ & RR $\uparrow$ & RRE $\downarrow$ & RTE $\downarrow$ & RR $\uparrow$ \\
         \midrule
         32 & \textbf{1.2} & \underline{5.5} & \underline{96.6} & \textbf{1.5} & \underline{6.6} & 95.6 & 3.2 & 12.2 & 72.8 & 2.9 & \underline{12.7} & \underline{72.0} \\
         48 & \textbf{1.2} & \textbf{5.4} & 96.4 & \textbf{1.5} & 6.9 & \underline{95.8} & \underline{3.1} & \textbf{11.9} & 74.2 & \textbf{2.6} & 13.6 & 69.0 \\
         64 (Ours) & \textbf{1.2} & \textbf{5.4} & \textbf{97.4} & \textbf{1.5} & 7.1 & \textbf{96.6} & \textbf{3.0} & 12.2 & \textbf{75.6} & \underline{2.8} & \textbf{12.3} & \textbf{73.6} \\
         96 & \underline{1.3} & 5.6 & 96.0 & \textbf{1.5} & \textbf{6.5} & 94.4 & \underline{3.1} & \underline{12.1} & \underline{74.4} & 3.1 & 14.6 & 69.2 \\
         \bottomrule
    \end{tabular}}
    \caption{Ablation studies of our \ourmethod{} with different numbers of correspondence in Top-K selection.}
    \label{tab: ablation_superpoint_topk}
\end{table*}

%% file: tables/ablation_RANSAC.tex
\begin{table*}[t]
    \setlength{\tabcolsep}{3pt}
    \centering
    \resizebox{\linewidth}{!}{
    \begin{tabular}{l|ccccc|ccccc|ccccc|ccccc}
        \toprule
        \multirow{4}{*}{\# Samples} & \multicolumn{10}{c|}{HybridMatch} & \multicolumn{10}{c}{HybridLoMatch} \\
        \cmidrule{2-21}
         & \multicolumn{5}{c|}{\textit{10-30\% non-rigid}} & \multicolumn{5}{c|}{\textit{30-50\% non-rigid}} & \multicolumn{5}{c|}{\textit{10-30\% non-rigid}} & \multicolumn{5}{c}{\textit{30-50\% non-rigid}}\\
        \cmidrule{2-21}
        & 5000 & 2500 & 1000 & 500 & 250 & 5000 & 2500 & 1000 & 500 & 250 & 5000 & 2500 & 1000 & 500 & 250 & 5000 & 2500 & 1000 & 500 & 250 \\
        \midrule
        \multicolumn{21}{c}{Feature Matching Recall (\%) $\uparrow$} \\
        \midrule
        (a) Baseline (GeoTrans) & \textbf{99.9} & \textbf{99.8} & \textbf{99.8} & \textbf{99.8} & \textbf{99.8} & \underline{99.4} & \underline{99.4} & \textbf{99.6} & \underline{99.4} & \underline{99.4} & \underline{93.2} & \underline{92.8} & \underline{93.4} & 92.2 & 92.0 & 91.4 & 91.4 & 90.4 & 91.6 & 90.8\\
        (b) w/ Uncertainty Mask + L1 Loss & 99.4 &\underline{ 99.2} & \underline{99.4} & \underline{99.6} & \underline{99.6} & \underline{99.4} & \underline{99.4} & \underline{99.2} & \underline{99.4} & \underline{99.4} & \textbf{94.0} & \textbf{93.8} & \textbf{94.2} & \textbf{93.6} & \textbf{93.8} & \underline{94.0} & \underline{94.0} & 93.8 & 93.4 & 93.2\\
        (c) w/ Uncertainty Mask + NLL Loss & \underline{99.8} & \textbf{99.8} & \textbf{99.8} & \textbf{99.8} & \textbf{99.8} & \textbf{99.6} & \textbf{99.6} & \textbf{99.6} & \textbf{99.6} & \underline{99.4} & 92.8 & \underline{92.8} & 92.6 & \underline{93.2} & 92.6 & \textbf{94.4} & \textbf{94.4} & \underline{94.2} & \underline{94.4} & \underline{94.4}\\
        (d) + Uncertainty-aware Correlation & \underline{99.8} & \textbf{99.8} & \textbf{99.8} & \textbf{99.8} & \textbf{99.8} & \textbf{99.6} & \textbf{99.6} & \textbf{99.6} & \textbf{99.6} & \textbf{99.6} & 92.8 & \underline{92.8} & 92.8 & 92.6 & \underline{93.0} & \textbf{94.4} & \textbf{94.4} & \textbf{94.8} & \textbf{94.6} & \textbf{94.8}\\

        \midrule
        \multicolumn{21}{c}{Inlier Ratio (\%) $\uparrow$} \\
        \midrule
        (a) Baseline (GeoTrans) & 36.5 & 36.5 & 36.4 & 36.7 & 36.4 & 33.6 & 33.6 & 33.7 & 33.6 & 33.6 & 18.2 & 18.3 & 18.3 & 18.3 & 18.4 & \underline{33.6} & 17.3 & 17.4 & 17.3 & 17.2 \\

        (b) w/ Uncertainty Mask + L1 Loss & \underline{52.8} & \underline{52.8} & 52.8 & \underline{52.9} & 53.1 & \underline{48.9} & \underline{48.9} & \underline{48.8} & \underline{48.9} & 48.9 & 29.1 & \underline{29.2} & \underline{29.2} & \underline{29.1} & \underline{29.2} & 27.8 & \underline{27.8} & 27.8 & 27.8 & \underline{27.7} \\
        
        (c) w/ Uncertainty Mask + NLL Loss & \textbf{58.6} & \textbf{58.7} & \underline{58.6} & \textbf{58.8} & \underline{58.5} & \textbf{54.4} & \textbf{54.4} & \textbf{54.4} & \textbf{54.3} & \textbf{54.4} & \underline{34.3} & \textbf{34.3} & \textbf{34.3} & \textbf{34.3} & \textbf{34.3} & \textbf{33.8} & \textbf{33.9} & \textbf{33.8} & \textbf{33.8} & \textbf{33.7} \\
        
        (d) + Uncertainty-aware Correlation & \textbf{58.6} & \textbf{58.7} & \textbf{58.7} & \textbf{58.8} & \textbf{58.6} & \textbf{54.4} & \textbf{54.4} & \textbf{54.4} & \textbf{54.3} & \underline{54.3} & \textbf{34.4} & \textbf{34.3} & \textbf{34.3} & \textbf{34.3} & \textbf{34.3} & \textbf{33.8} & \textbf{33.9} & \underline{33.7} & \underline{33.7} & \textbf{33.7} \\
        
        
        

        \midrule
        \multicolumn{21}{c}{Registration Recall (\%) $\uparrow$} \\
        \midrule
        (a) Baseline (GeoTrans) & 92.8 & 91.4 & 89.0 & 88.6 & 90.0 & 91.0 & 86.8 & 88.6 & 86.8 & 86.4 & 52.6 & 51.4 & 54.4 & 49.2 & 51.8 & 52.4 & 55.2 & 43.8 & 50.0 & 50.8 \\

        (b) w/ Uncertainty Mask + L1 Loss & 93.2 & 93.4 & 94.0 & \underline{94.0} & 92.6 & 94.8 & 95.1 & 95.0 & 93.8 & 94.2 & 62.4 & 64.2 & \underline{66.0} & 64.6 & 63.0 & \underline{61.0} & \underline{61.6} & 61.4 & 60.8 & 61.0 \\
        
        (c) w/ Uncertainty Mask + NLL Loss & \underline{97.2} & \textbf{96.8} & \underline{97.0} & \textbf{97.2} & \textbf{96.8} & \textbf{96.8} & \underline{96.5} & \underline{96.6} & \textbf{95.2} & \textbf{96.4} & \underline{73.4} & \textbf{73.6} & \textbf{75.2} & \underline{72.2} & \textbf{73.2} & \underline{76.0} & \textbf{75.8} & \textbf{75.4} & \underline{71.4} & \textbf{73.2} \\
        
        (d) + Uncertainty-aware Correlation & \textbf{97.4} & \underline{96.4} & \textbf{97.6} & \textbf{97.2} & \underline{96.4} & \underline{96.6} & \textbf{97.4} & \textbf{97.8} & \underline{95.0} & \underline{96.0} & \textbf{75.6} & \underline{73.0} & \textbf{75.2} & \textbf{73.0} &\underline{70.8} & \textbf{76.2} & \textbf{75.8} & \underline{75.0} & \textbf{74.2} & \underline{72.4} \\
        
        
        

        \bottomrule
    \end{tabular}}
    \caption{Ablation study of our \ourmethod{} using RANSAC as the transformation estimator. }
    \label{tab: ablation_RANSAC}
\end{table*}

%% file: camera_ready.bbl
\begin{thebibliography}{71}
\providecommand{\natexlab}[1]{#1}

\bibitem[{Ao et~al.(2023)Ao, Hu, Wang, Xu, and Guo}]{Ao_2023_CVPR}
Ao, S.; Hu, Q.; Wang, H.; Xu, K.; and Guo, Y. 2023.
\newblock {BUFFER}: Balancing Accuracy, Efficiency, and Generalizability in Point Cloud Registration.
\newblock In \emph{CVPR}, 1255--1264.

\bibitem[{Ao et~al.(2021)Ao, Hu, Yang, Markham, and Guo}]{ao2021spinnet}
Ao, S.; Hu, Q.; Yang, B.; Markham, A.; and Guo, Y. 2021.
\newblock {SpinNet}: Learning a General Surface Descriptor for {3D} Point Cloud Registration.
\newblock In \emph{CVPR}, 11753--11762.

\bibitem[{Aoki et~al.(2019)Aoki, Goforth, Srivatsan, and Lucey}]{aoki2019pointnetlk}
Aoki, Y.; Goforth, H.; Srivatsan, R.~A.; and Lucey, S. 2019.
\newblock {PointNetLK}: Robust \& Efficient Point Cloud Registration using {PointNet}.
\newblock In \emph{CVPR}, 7163--7172.

\bibitem[{Bai et~al.(2021)Bai, Luo, Zhou, Chen, Li, Hu, Fu, and Tai}]{bai2021pointdsc}
Bai, X.; Luo, Z.; Zhou, L.; Chen, H.; Li, L.; Hu, Z.; Fu, H.; and Tai, C.-L. 2021.
\newblock {PointDSC}: Robust Point Cloud Registration Using Deep Spatial Consistency.
\newblock In \emph{CVPR}, 15859--15869.

\bibitem[{Bai et~al.(2020)Bai, Luo, Zhou, Fu, Quan, and Tai}]{bai2020d3feat}
Bai, X.; Luo, Z.; Zhou, L.; Fu, H.; Quan, L.; and Tai, C.-L. 2020.
\newblock {D3Feat}: Joint Learning of Dense Detection and Description of {3D} Local Features.
\newblock In \emph{CVPR}, 6359--6367.

\bibitem[{Besl and McKay(1992)}]{besl1992method}
Besl, P.~J.; and McKay, N.~D. 1992.
\newblock A Method for Registration of {3D} Shapes.
\newblock \emph{IEEE TPAMI}, 14(2): 239--256.

\bibitem[{Bozic et~al.(2020{\natexlab{a}})Bozic, Palafox, Zollh{\"o}fer, Dai, Thies, and Nie{\ss}ner}]{bozic2020neural}
Bozic, A.; Palafox, P.; Zollh{\"o}fer, M.; Dai, A.; Thies, J.; and Nie{\ss}ner, M. 2020{\natexlab{a}}.
\newblock Neural non-rigid tracking.
\newblock \emph{NeurIPS}, 33: 18727--18737.

\bibitem[{Bozic et~al.(2020{\natexlab{b}})Bozic, Zollhofer, Theobalt, and Nie{\ss}ner}]{bozic2020deepdeform}
Bozic, A.; Zollhofer, M.; Theobalt, C.; and Nie{\ss}ner, M. 2020{\natexlab{b}}.
\newblock {DeepDeform}: Learning non-rigid {RGB-D} reconstruction with semi-supervised data.
\newblock In \emph{CVPR}, 7002--7012.

\bibitem[{Chang et~al.(2015)Chang, Funkhouser, Guibas, Hanrahan, Huang, Li, Savarese, Savva, Song, Su et~al.}]{chang2015shapenet}
Chang, A.~X.; Funkhouser, T.; Guibas, L.~J.; Hanrahan, P.; Huang, Q.; Li, Z.; Savarese, S.; Savva, M.; Song, S.; Su, H.; et~al. 2015.
\newblock {ShapeNet}: An Information-Rich {3D} Model Repository.
\newblock \emph{arXiv preprint arXiv:1512.03012}.

\bibitem[{Chen et~al.(2023{\natexlab{a}})Chen, Wang, Yuan, Yang, and Yue}]{Chen_2023_ICCV}
Chen, G.; Wang, M.; Yuan, L.; Yang, Y.; and Yue, Y. 2023{\natexlab{a}}.
\newblock Rethinking Point Cloud Registration as Masking and Reconstruction.
\newblock In \emph{ICCV}, 17717--17727.

\bibitem[{Chen et~al.(2024)Chen, Xu, Li, Luo, Liu, Fu, Tan, and Liu}]{chen2024pointreggpt}
Chen, S.; Xu, H.; Li, H.; Luo, K.; Liu, G.; Fu, C.-W.; Tan, P.; and Liu, S. 2024.
\newblock {PointRegGPT}: Boosting {3D} Point Cloud Registration using Generative Point-Cloud Pairs for Training.
\newblock In \emph{ECCV}.

\bibitem[{Chen et~al.(2023{\natexlab{b}})Chen, Xu, Li, Liu, Fu, and Liu}]{chen2023sira}
Chen, S.; Xu, H.; Li, R.; Liu, G.; Fu, C.-W.; and Liu, S. 2023{\natexlab{b}}.
\newblock {SIRA-PCR}: Sim-to-Real Adaptation for {3D} Point Cloud Registration.
\newblock In \emph{ICCV}, 14394--14405.

\bibitem[{Choy, Dong, and Koltun(2020)}]{choy2020deep}
Choy, C.; Dong, W.; and Koltun, V. 2020.
\newblock Deep Global Registration.
\newblock In \emph{CVPR}, 2514--2523.

\bibitem[{Choy, Park, and Koltun(2019)}]{choy2019fully}
Choy, C.; Park, J.; and Koltun, V. 2019.
\newblock Fully Convolutional Geometric Features.
\newblock In \emph{ICCV}, 8958--8966.

\bibitem[{Curless and Levoy(1996)}]{curless1996volumetric}
Curless, B.; and Levoy, M. 1996.
\newblock A Volumetric Method for Building Complex Models from Range Images.
\newblock In \emph{ACM SIGGRAPH}, 303--312.

\bibitem[{Dang and Salzmann(2023)}]{Dang_2023_ICCV}
Dang, Z.; and Salzmann, M. 2023.
\newblock {AutoSynth}: Learning to Generate {3D} Training Data for Object Point Cloud Registration.
\newblock In \emph{ICCV}, 9009--9019.

\bibitem[{Dosovitskiy et~al.(2015)Dosovitskiy, Fischer, Ilg, Hausser, Hazirbas, Golkov, Van Der~Smagt, Cremers, and Brox}]{dosovitskiy2015flownet}
Dosovitskiy, A.; Fischer, P.; Ilg, E.; Hausser, P.; Hazirbas, C.; Golkov, V.; Van Der~Smagt, P.; Cremers, D.; and Brox, T. 2015.
\newblock {FlowNet}: Learning Optical Flow With Convolutional Networks.
\newblock In \emph{ICCV}, 2758--2766.

\bibitem[{Fischler and Bolles(1981)}]{fischler1981random}
Fischler, M.~A.; and Bolles, R.~C. 1981.
\newblock Random Sample Consensus: A Paradigm for Model Fitting With Applications to Image Analysis and Automated Cartography.
\newblock \emph{Communications of the ACM}, 24(6): 381--395.

\bibitem[{Fu et~al.(2021{\natexlab{a}})Fu, Cai, Gao, Zhang, Wang, Li, Zeng, Sun, Jia, Zhao et~al.}]{fu20213d}
Fu, H.; Cai, B.; Gao, L.; Zhang, L.-X.; Wang, J.; Li, C.; Zeng, Q.; Sun, C.; Jia, R.; Zhao, B.; et~al. 2021{\natexlab{a}}.
\newblock {3D-FRONT}: {3D} Furnished Rooms with Layouts and Semantics.
\newblock In \emph{ICCV}, 10933--10942.

\bibitem[{Fu et~al.(2021{\natexlab{b}})Fu, Liu, Luo, and Wang}]{fu2021robust}
Fu, K.; Liu, S.; Luo, X.; and Wang, M. 2021{\natexlab{b}}.
\newblock Robust Point Cloud Registration Framework Based on Deep Graph Matching.
\newblock In \emph{CVPR}, 8893--8902.

\bibitem[{Gao and Tedrake(2018)}]{gao18surfelwarp}
Gao, W.; and Tedrake, R. 2018.
\newblock {SurfelWarp}: Efficient Non-Volumetric Single View Dynamic Reconstruction.
\newblock In \emph{RSS}.

\bibitem[{Gojcic et~al.(2019)Gojcic, Zhou, Wegner, and Wieser}]{gojcic2019perfect}
Gojcic, Z.; Zhou, C.; Wegner, J.~D.; and Wieser, A. 2019.
\newblock The Perfect Match: {3D} Point Cloud Matching with Smoothed Densities.
\newblock In \emph{CVPR}, 5545--5554.

\bibitem[{Groueix et~al.(2018)Groueix, Fisher, Kim, Russell, and Aubry}]{groueix20183d}
Groueix, T.; Fisher, M.; Kim, V.~G.; Russell, B.~C.; and Aubry, M. 2018.
\newblock {3D-CODED}: {3D} correspondences by deep deformation.
\newblock In \emph{ECCV}, 230--246.

\bibitem[{Hatem, Qian, and Wang(2023)}]{Hatem_2023_ICCV}
Hatem, A.; Qian, Y.; and Wang, Y. 2023.
\newblock {Point-TTA}: Test-Time Adaptation for Point Cloud Registration Using Multitask Meta-Auxiliary Learning.
\newblock In \emph{ICCV}, 16494--16504.

\bibitem[{Huang et~al.(2008)Huang, Adams, Wicke, and Guibas}]{huang2008non}
Huang, Q.-X.; Adams, B.; Wicke, M.; and Guibas, L.~J. 2008.
\newblock Non-rigid registration under isometric deformations.
\newblock In \emph{Comput. Graph. Forum}, volume~27, 1449--1457.

\bibitem[{Huang et~al.(2021)Huang, Gojcic, Usvyatsov, Wieser, and Schindler}]{huang2020predator}
Huang, S.; Gojcic, Z.; Usvyatsov, M.; Wieser, A.; and Schindler, K. 2021.
\newblock {PREDATOR}: Registration of {3D} Point Clouds with Low Overlap.
\newblock In \emph{CVPR}, 4267--4276.

\bibitem[{Ilg et~al.(2018)Ilg, Cicek, Galesso, Klein, Makansi, Hutter, and Brox}]{ilg2018uncertainty}
Ilg, E.; Cicek, O.; Galesso, S.; Klein, A.; Makansi, O.; Hutter, F.; and Brox, T. 2018.
\newblock Uncertainty estimates and multi-hypotheses networks for optical flow.
\newblock In \emph{ECCV}, 652--667.

\bibitem[{Innmann et~al.(2016)Innmann, Zollh{\"o}fer, Nie{\ss}ner, Theobalt, and Stamminger}]{innmann2016volumedeform}
Innmann, M.; Zollh{\"o}fer, M.; Nie{\ss}ner, M.; Theobalt, C.; and Stamminger, M. 2016.
\newblock {VolumeDeform}: Real-time volumetric non-rigid reconstruction.
\newblock In \emph{ECCV}, 362--379.

\bibitem[{Jiang et~al.(2023)Jiang, Dang, Wei, Xie, Yang, and Salzmann}]{Jiang_2023_CVPR}
Jiang, H.; Dang, Z.; Wei, Z.; Xie, J.; Yang, J.; and Salzmann, M. 2023.
\newblock Robust Outlier Rejection for {3D} Registration With Variational Bayes.
\newblock In \emph{CVPR}, 1148--1157.

\bibitem[{Kendall and Gal(2017)}]{kendall2017uncertainties}
Kendall, A.; and Gal, Y. 2017.
\newblock What uncertainties do we need in bayesian deep learning for computer vision?
\newblock \emph{NeurIPS}, 30.

\bibitem[{Kingma and Ba(2015)}]{kingma2014adam}
Kingma, P.~D.; and Ba, L.~J. 2015.
\newblock Adam: A Method for Stochastic Optimization.
\newblock In \emph{ICLR}.

\bibitem[{Li et~al.(2020{\natexlab{a}})Li, Zhang, Xu, Zhou, and Zhang}]{idam}
Li, J.; Zhang, C.; Xu, Z.; Zhou, H.; and Zhang, C. 2020{\natexlab{a}}.
\newblock Iterative Distance-Aware Similarity Matrix Convolution with Mutual-Supervised Point Elimination for Efficient Point Cloud Registration.
\newblock In \emph{ECCV}, 378--394.

\bibitem[{Li, Kaesemodel~Pontes, and Lucey(2021)}]{li2021neural}
Li, X.; Kaesemodel~Pontes, J.; and Lucey, S. 2021.
\newblock Neural scene flow prior.
\newblock \emph{NeurIPS}, 34: 7838--7851.

\bibitem[{Li et~al.(2020{\natexlab{b}})Li, Bozic, Zhang, Ji, Harada, and Nie{\ss}ner}]{li2020learning}
Li, Y.; Bozic, A.; Zhang, T.; Ji, Y.; Harada, T.; and Nie{\ss}ner, M. 2020{\natexlab{b}}.
\newblock Learning to optimize non-rigid tracking.
\newblock In \emph{CVPR}, 4910--4918.

\bibitem[{Li and Harada(2022)}]{lepard2021}
Li, Y.; and Harada, T. 2022.
\newblock Lepard: Learning Partial Point Cloud Matching in Rigid and Deformable Scenes.
\newblock \emph{CVPR}.

\bibitem[{Li et~al.(2021)Li, Takehara, Taketomi, Zheng, and Nie{\ss}ner}]{li20214dcomplete}
Li, Y.; Takehara, H.; Taketomi, T.; Zheng, B.; and Nie{\ss}ner, M. 2021.
\newblock {4Dcomplete}: Non-rigid motion estimation beyond the observable surface.
\newblock In \emph{ICCV}, 12706--12716.

\bibitem[{Liu et~al.(2023{\natexlab{a}})Liu, Wang, Liu, Jiang, Pollefeys, and Wang}]{liu2023regformer}
Liu, J.; Wang, G.; Liu, Z.; Jiang, C.; Pollefeys, M.; and Wang, H. 2023{\natexlab{a}}.
\newblock {RegFormer}: An efficient projection-aware transformer network for large-scale point cloud registration.
\newblock In \emph{ICCV}, 8451--8460.

\bibitem[{Liu et~al.(2023{\natexlab{b}})Liu, Zhu, Zhou, Li, Chang, and Guo}]{Liu_2023_ICCV}
Liu, Q.; Zhu, H.; Zhou, Y.; Li, H.; Chang, S.; and Guo, M. 2023{\natexlab{b}}.
\newblock Density-invariant Features for Distant Point Cloud Registration.
\newblock In \emph{ICCV}, 18215--18225.

\bibitem[{Liu, Qi, and Guibas(2019)}]{liu2019flownet3d}
Liu, X.; Qi, C.~R.; and Guibas, L.~J. 2019.
\newblock {FlowNet3D}: Learning scene flow in {3D} point clouds.
\newblock In \emph{CVPR}, 529--537.

\bibitem[{Lowe(2004)}]{lowe2004distinctive}
Lowe, D.~G. 2004.
\newblock Distinctive image features from scale-invariant keypoints.
\newblock \emph{IJCV}, 60: 91--110.

\bibitem[{Mei et~al.(2023)Mei, Tang, Huang, Wang, Liu, Zhang, Van~Gool, and Wu}]{Mei_2023_CVPR}
Mei, G.; Tang, H.; Huang, X.; Wang, W.; Liu, J.; Zhang, J.; Van~Gool, L.; and Wu, Q. 2023.
\newblock Unsupervised Deep Probabilistic Approach for Partial Point Cloud Registration.
\newblock In \emph{CVPR}, 13611--13620.

\bibitem[{Newcombe, Fox, and Seitz(2015)}]{newcombe2015dynamicfusion}
Newcombe, R.~A.; Fox, D.; and Seitz, S.~M. 2015.
\newblock {DynamicFusion}: Reconstruction and tracking of non-rigid scenes in real-time.
\newblock In \emph{CVPR}, 343--352.

\bibitem[{Ovsjanikov et~al.(2012)Ovsjanikov, Ben-Chen, Solomon, Butscher, and Guibas}]{ovsjanikov2012functional}
Ovsjanikov, M.; Ben-Chen, M.; Solomon, J.; Butscher, A.; and Guibas, L. 2012.
\newblock Functional maps: a flexible representation of maps between shapes.
\newblock \emph{ACM TOG}, 31(4): 1--11.

\bibitem[{Pais et~al.(2020)Pais, Ramalingam, Govindu, Nascimento, Chellappa, and Miraldo}]{pais20203dregnet}
Pais, G.~D.; Ramalingam, S.; Govindu, V.~M.; Nascimento, J.~C.; Chellappa, R.; and Miraldo, P. 2020.
\newblock {3DRegNet}: A Deep Neural Network for {3D} Point Registration.
\newblock In \emph{CVPR}, 7193--7203.

\bibitem[{Poiesi and Boscaini(2022)}]{poiesi2022learning}
Poiesi, F.; and Boscaini, D. 2022.
\newblock Learning General and Distinctive {3D} Local Deep Descriptors for Point Cloud Registration.
\newblock \emph{IEEE TPAMI}.

\bibitem[{Pomerleau et~al.(2012)Pomerleau, Liu, Colas, and Siegwart}]{pomerleau2012challenging}
Pomerleau, F.; Liu, M.; Colas, F.; and Siegwart, R. 2012.
\newblock Challenging Data Sets for Point Cloud Registration Algorithms.
\newblock \emph{IJRR}, 31(14): 1705--1711.

\bibitem[{Qin et~al.(2022)Qin, Yu, Wang, Guo, Peng, and Xu}]{qin2022geometric}
Qin, Z.; Yu, H.; Wang, C.; Guo, Y.; Peng, Y.; and Xu, K. 2022.
\newblock Geometric Transformer for Fast and Robust Point Cloud Registration.
\newblock In \emph{CVPR}, 11143--11152.

\bibitem[{Rocco et~al.(2018)Rocco, Cimpoi, Arandjelovi{\'c}, Torii, Pajdla, and Sivic}]{rocco2018neighbourhood}
Rocco, I.; Cimpoi, M.; Arandjelovi{\'c}, R.; Torii, A.; Pajdla, T.; and Sivic, J. 2018.
\newblock Neighbourhood consensus networks.
\newblock \emph{NeurIPS}, 31.

\bibitem[{Rusinkiewicz(2019)}]{rusinkiewicz2019symmetric}
Rusinkiewicz, S. 2019.
\newblock A Symmetric Objective Function for {LiDAR}.
\newblock \emph{ACM TOG}, 38(4): 1--7.

\bibitem[{Rusinkiewicz and Levoy(2001)}]{rusinkiewicz-normal-sampling}
Rusinkiewicz, S.; and Levoy, M. 2001.
\newblock Efficient Variants of the {ICP} Algorithm.
\newblock In \emph{3DIM}, 145--152.

\bibitem[{Rusu, Blodow, and Beetz(2009)}]{FPFH}
Rusu, R.~B.; Blodow, N.; and Beetz, M. 2009.
\newblock Fast Point Feature Histograms ({FPFH}) for {3D} Registration.
\newblock In \emph{ICRA}, 3212--3217.

\bibitem[{Schmidt, Newcombe, and Fox(2016)}]{schmidt2016self}
Schmidt, T.; Newcombe, R.; and Fox, D. 2016.
\newblock Self-supervised visual descriptor learning for dense correspondence.
\newblock \emph{IEEE RAL}, 2(2): 420--427.

\bibitem[{Shotton et~al.(2013)Shotton, Glocker, Zach, Izadi, Criminisi, and Fitzgibbon}]{shotton2013scene}
Shotton, J.; Glocker, B.; Zach, C.; Izadi, S.; Criminisi, A.; and Fitzgibbon, A. 2013.
\newblock Scene Coordinate Regression Forests for Camera Relocalization in {RGB-D} Images.
\newblock In \emph{CVPR}, 2930--2937.

\bibitem[{Sun et~al.(2021)Sun, Shen, Wang, Bao, and Zhou}]{sun2021loftr}
Sun, J.; Shen, Z.; Wang, Y.; Bao, H.; and Zhou, X. 2021.
\newblock {LoFTR}: Detector-free local feature matching with transformers.
\newblock In \emph{CVPR}, 8922--8931.

\bibitem[{Thomas et~al.(2019)Thomas, Qi, Deschaud, Marcotegui, Goulette, and Guibas}]{thomas2019kpconv}
Thomas, H.; Qi, C.~R.; Deschaud, J.-E.; Marcotegui, B.; Goulette, F.; and Guibas, L.~J. 2019.
\newblock {KPConv}: Flexible and Deformable Convolution for Point Clouds.
\newblock In \emph{ICCV}, 6411--6420.

\bibitem[{Truong et~al.(2023)Truong, Danelljan, Timofte, and Van~Gool}]{truong2023pdc}
Truong, P.; Danelljan, M.; Timofte, R.; and Van~Gool, L. 2023.
\newblock {PDC-Net+}: Enhanced probabilistic dense correspondence network.
\newblock \emph{IEEE TPAMI}, 45(8): 10247--10266.

\bibitem[{Wang et~al.(2022)Wang, Liu, Dong, and Wang}]{wang2022you}
Wang, H.; Liu, Y.; Dong, Z.; and Wang, W. 2022.
\newblock You Only Hypothesize Once: Point Cloud Registration With Rotation-Equivariant Descriptors.
\newblock In \emph{IEEE TMM}, 1630--1641.

\bibitem[{Wang et~al.(2023)Wang, Liu, Hu, Wang, Chen, Dong, Guo, Wang, and Yang}]{wang2023roreg}
Wang, H.; Liu, Y.; Hu, Q.; Wang, B.; Chen, J.; Dong, Z.; Guo, Y.; Wang, W.; and Yang, B. 2023.
\newblock {RoReg}: Pairwise Point Cloud Registration with Oriented Descriptors and Local Rotations.
\newblock \emph{IEEE TPAMI}.

\bibitem[{Wang and Solomon(2019{\natexlab{a}})}]{wang2019deep}
Wang, Y.; and Solomon, J.~M. 2019{\natexlab{a}}.
\newblock Deep Closest Point: Learning Representations for Point Cloud Registration.
\newblock In \emph{ICCV}, 3523--3532.

\bibitem[{Wang and Solomon(2019{\natexlab{b}})}]{wang2019prnet}
Wang, Y.; and Solomon, J.~M. 2019{\natexlab{b}}.
\newblock {PRNet}: Self-Supervised Learning for Partial-to-Partial Registration.
\newblock In \emph{NeurIPS}, 8814--8826.

\bibitem[{Wu et~al.(2020)Wu, Wang, Li, Liu, and Fuxin}]{wu2020pointpwc}
Wu, W.; Wang, Z.~Y.; Li, Z.; Liu, W.; and Fuxin, L. 2020.
\newblock PointPWC-Net: Cost Volume on Point Clouds for (Self-) Supervised Scene Flow Estimation.
\newblock In \emph{ECCV}, 88--107.

\bibitem[{Xu et~al.(2021)Xu, Liu, Wang, Liu, and Zeng}]{xu2021omnet}
Xu, H.; Liu, S.; Wang, G.; Liu, G.; and Zeng, B. 2021.
\newblock {OMNet}: Learning Overlapping Mask for Partial-to-Partial Point Cloud Registration.
\newblock In \emph{ICCV}.

\bibitem[{Xu et~al.(2022)Xu, Ye, Liu, Zeng, and Liu}]{xu2022finet}
Xu, H.; Ye, N.; Liu, G.; Zeng, B.; and Liu, S. 2022.
\newblock {FINet}: Dual Branches Feature Interaction for Partial-To-Partial Point Cloud Registration.
\newblock In \emph{AAAI}, volume~36, 2848--2856.

\bibitem[{Yang et~al.(2022)Yang, Guo, Chen, and Tao}]{yangone}
Yang, F.; Guo, L.; Chen, Z.; and Tao, W. 2022.
\newblock One-inlier is first: Towards efficient position encoding for point cloud registration.
\newblock \emph{NeurIPS}, 35: 6982--6995.

\bibitem[{Yang, Li, and Jia(2013)}]{yang2013go}
Yang, J.; Li, H.; and Jia, Y. 2013.
\newblock {Go-ICP}: Solving {3D} Registration Efficiently and Globally Optimally.
\newblock In \emph{CVPR}, 1457--1464.

\bibitem[{Yew and Lee(2020)}]{yew2020-RPMNet}
Yew, Z.~J.; and Lee, G.~H. 2020.
\newblock {RPM-Net}: Robust Point Matching using Learned Features.
\newblock In \emph{CVPR}, 11824--11833.

\bibitem[{Yew and Lee(2022)}]{yew2022regtr}
Yew, Z.~J.; and Lee, G.~H. 2022.
\newblock {REGTR}: End-to-end Point Cloud Correspondences with Transformers.
\newblock In \emph{CVPR}, 6677--6686.

\bibitem[{Yu et~al.(2021)Yu, Li, Saleh, Busam, and Ilic}]{yu2021cofinet}
Yu, H.; Li, F.; Saleh, M.; Busam, B.; and Ilic, S. 2021.
\newblock {CoFiNet}: Reliable Coarse-to-Fine Correspondences for Robust Point Cloud Registration.
\newblock \emph{NeurIPS}, 34: 23872--23884.

\bibitem[{Yu et~al.(2023)Yu, Ren, Zhang, Zhou, Lin, and Dai}]{yu2023peal}
Yu, J.; Ren, L.; Zhang, Y.; Zhou, W.; Lin, L.; and Dai, G. 2023.
\newblock {PEAL}: Prior-Embedded Explicit Attention Learning for Low-Overlap Point Cloud Registration.
\newblock In \emph{CVPR}, 17702--17711.

\bibitem[{Zeng et~al.(2017)Zeng, Song, Nie{\ss}ner, Fisher, Xiao, and Funkhouser}]{zeng20173dmatch}
Zeng, A.; Song, S.; Nie{\ss}ner, M.; Fisher, M.; Xiao, J.; and Funkhouser, T. 2017.
\newblock {3DMatch}: Learning Local Geometric Descriptors from {RGB-D} Reconstructions.
\newblock In \emph{CVPR}, 1802--1811.

\bibitem[{Zhang et~al.(2023)Zhang, Yang, Zhang, and Zhang}]{Zhang_2023_CVPR}
Zhang, X.; Yang, J.; Zhang, S.; and Zhang, Y. 2023.
\newblock {3D} Registration With Maximal Cliques.
\newblock In \emph{CVPR}, 17745--17754.

\end{thebibliography}
